%% file: main.tex
\documentclass[sigconf]{acmart}
\AtBeginDocument{%
  }

\usepackage{xcolor}

\usepackage{pdflscape}

\usepackage{tikz}
\usepackage{subcaption}
\usepackage{caption}

\usepackage[normalem]{ulem}

\usepackage{fontspec}
\usepackage{ucharclasses}

\newfontfamily\cjkfont{NotoSansCJK-Regular.ttc}[Path=./]
\setTransitionsForCJK{\cjkfont}{\ttfamily}

\newfontfamily\devanagarifont{NotoSansDevanagari-VariableFont_wdth,wght.ttf}[Path=./]
\setTransitionsForDevanagari{\devanagarifont}{\ttfamily}

\usepackage{twemojis}
\newfontfamily{\cyrillicfont}{NotoSans-VariableFont_wdth,wght.ttf}[Path=./, Script=Cyrillic]
\setTransitionsForCyrillics{\cyrillicfont}{\normalfont}

\newfontfamily{\arabicfont}{NotoSansArabic-VariableFont_wdth,wght.ttf}[Path=./]
\setTransitionsFor{Arabic}{\arabicfont}{\normalfont}
\setTransitionsFor{ArabicSupplement}{\arabicfont}{\normalfont}
\setTransitionsFor{ArabicExtendedA}{\arabicfont}{\normalfont}

\newfontfamily{\chakmafont}{NotoSansChakma-Regular.ttf}[Path=./]
\setTransitionsFor{Chakma}{\chakmafont}{\normalfont}
\setTransitionsFor{ChakmaExtended}{\chakmafont}{\normalfont}

\newfontfamily{\bamumfont}{NotoSansBamum-VariableFont_wght.ttf}[Path=./]
\setTransitionsFor{Bamum}{\bamumfont}{\normalfont}
\setTransitionsFor{BamumSupplement}{\bamumfont}{\normalfont}

\newfontfamily{\hebrewfont}{NotoSansHebrew-Regular.ttf}[Path=./]
\setTransitionsFor{Hebrew}{\hebrewfont}{\normalfont}

\newfontfamily{\musicfont}{NotoMusic-Regular.ttf}[Path=./]
\setTransitionsFor{MusicalSymbols}{\musicfont}{\normalfont}

\setcopyright{acmlicensed}
\copyrightyear{2025}
\acmYear{2025}
\acmDOI{10.1145/3715275.3732068}
\acmConference[FAccT '25]{The 2025 ACM Conference on Fairness, Accountability, and Transparency}{June 23--26, 2025}{Athens, Greece}

\begin{document}

\title{Reward Model Interpretability via Optimal and Pessimal Tokens}

\author{Brian Christian}
\email{brian.christian@psy.ox.ac.uk}
\orcid{0000-0001-5277-8939}
\affiliation{%
  \institution{University of Oxford}
  \city{Oxford}
  \country{UK}
}   
\author{Hannah Rose Kirk}
\email{hannah.kirk@oii.ox.ac.uk}
\orcid{0000-0002-7419-5993}
\affiliation{%
  \institution{University of Oxford}
  \city{Oxford}
  \country{UK}
}
\author{Jessica A.F. Thompson}
\email{jessica.thompson@psy.ox.ac.uk}
\orcid{0000-0003-0468-5097}
\affiliation{%
  \institution{University of Oxford}
  \city{Oxford}
  \country{UK}
}
\author{Christopher Summerfield}
\authornote{Both authors contributed equally to the paper}
\email{christopher.summerfield@psy.ox.ac.uk}
\orcid{0000-0002-2941-2653}
\affiliation{%
  \institution{University of Oxford}
  \city{Oxford}
  \country{UK} 
}
\author{Tsvetomira Dumbalska}
\authornotemark[1]
\email{tsvetomira.dumbalska@psy.ox.ac.uk}
\orcid{0000-0002-5761-8536}
\affiliation{%
  \institution{University of Oxford}
  \city{Oxford}
  \country{UK}
}

\renewcommand{\shortauthors}{B. Christian, H.R. Kirk, J.A.F. Thompson, C. Summerfield, T. Dumbalska}

\begin{abstract}
  \input{sections/abstract.tex}
\end{abstract}







\maketitle

\color{red}
CONTENT WARNING: This article presents examples of biased, offensive, sexually explicit and otherwise harmful text. The authors do not endorse any of the harmful representations quoted below.
\color{black}

\definecolor{N-Gem-27B}{rgb}{0.121568, 0.466666, 0.705882} 
\definecolor{S-Gem-27B-v0.2}{rgb}{1.0, 0.498039, 0.054901} 
\definecolor{S-Gem-27B}{rgb}{0.172549, 0.627450, 0.172549} 
\definecolor{S-Lla-8B-v0.2}{rgb}{0.839215, 0.152941, 0.156862} 
\definecolor{N-Lla-8B}{rgb}{0.580392, 0.403921, 0.741176} 
\definecolor{L-Lla-8B}{rgb}{0.549019, 0.337254, 0.294117} 
\definecolor{R-Lla-8B}{rgb}{0.890196, 0.466666, 0.760784} 
\definecolor{R-Lla-3B}{rgb}{0.498039, 0.498039, 0.498039} 
\definecolor{F-Lla-8B-v0.1}{rgb}{0.737254, 0.741176, 0.133333} 
\definecolor{R-Gem-2B}{rgb}{0.090196, 0.745098, 0.811764} 

\newcommand{\modelcolor}[1]{%
  \textcolor{#1}{■}%
}
\newcommand{\modelname}[1]{%
  \modelcolor{#1}~#1%
}

\input{sections/introduction}
\input{sections/ranking-tokens.tex}

\input{sections/framing.tex}
\input{sections/comparison-to-humans.tex}
\input{sections/gcg.tex}
\input{sections/related-work}
\input{sections/conclusion.tex}

\begin{acks}
Thank you to Franziska Brändle, Owain Evans, Matan Mazor, and Carroll Wainwright for helpful discussions.
\end{acks}

\bibliographystyle{ACM-Reference-Format}
\bibliography{references}

\clearpage
\appendix
\onecolumn
\counterwithin{figure}{section}
\counterwithin{table}{section}
\input{supplement.tex}

\end{document}

%% file: sections/abstract.tex
Reward modeling has emerged as a crucial component in aligning large language models with human values. Significant attention has focused on using reward models as a means for fine-tuning generative models. However, the reward models themselves---which directly encode human value judgments by turning prompt-response pairs into scalar rewards---remain relatively understudied. We present a novel approach to reward model interpretability through exhaustive analysis of their responses across their entire vocabulary space. By examining how different reward models score every possible single-token response to value-laden prompts, we uncover several striking findings: (i) substantial heterogeneity between models trained on similar objectives, (ii) systematic asymmetries in how models encode high- vs low-scoring tokens, (iii) significant sensitivity to prompt framing that mirrors human cognitive biases, and (iv) overvaluation of more frequent tokens. We demonstrate these effects across ten recent open-source reward models of varying parameter counts and architectures. Our results challenge assumptions about the interchangeability of reward models, as well as their suitability as proxies of complex and context-dependent human values. We find that these models can encode concerning biases toward certain identity groups, which may emerge as unintended consequences of harmlessness training---distortions that risk propagating through the downstream large language models now deployed to millions.

%% file: sections/introduction.tex
\section{Introduction}

The alignment of large language models (LLMs) with human values has emerged as one of the central challenges in modern AI development, and at the heart of this challenge lie ``reward models''---neural networks trained to directly proxy human preferences by transforming text into scalar rewards. Though typically treated as disposable intermediaries in the larger alignment process, these models are crucial objects of study in their own right as the most direct and explicit encoding of human values in AI systems, yet are surprisingly under-explored.

The typical process for aligning an LLM with human values involves collecting a dataset of labeled pairwise human preferences, indicating which of two LLM responses to a given user prompt is preferred \cite{christiano2017deep}. These preference data often distill multiple desirable objectives, such as helpfulness, harmlessness, and honesty \cite{bai2022traininghelpfulharmlessassistant}, which are operationalized through guidelines written by model developers and interpreted by crowdworkers \cite{kirk2023past}. The resulting dataset is used to train a ``reward model''---a transformer model that takes in a prompt-response pair (or a longer user-assistant dialogue) and outputs a scalar that represents in effect how ``preferable'' that response is. These scalars are typically based on the Bradley-Terry score \cite{bradley1952rank}, and the reward model is trained via stochastic gradient descent to minimize the negative log-likelihood of the observed pairwise preferences \cite{lambert2023entangled}. The trained reward model then acts as a scalable proxy for human preferences when using reinforcement-learning algorithms such as Proximal Policy Optimization (PPO) \cite{schulman2017proximal} to fine-tune the LLM. This process---known as Reinforcement Learning from Human Feedback (RLHF)---results in LLM generations that maximize the reward model's score rather than the pre-training objective, so are supposedly more aligned with human values. While direct alignment algorithms like Direct Preference Optimization (DPO) \cite{rafailov2023direct} have grown in popularity, they capture equivalent preference relationships from the data, just without the reward model as an intermediary.

Although reward models exist as a disposable reagent in the process of turning an ``unaligned'' LLM built to minimize predictive loss into an ``aligned'' one built to maximize this proxy of human preference, they are fascinating objects of research inquiry in their own right. Designed as generalizable proxies for human preference, they offer more direct value encoding than downstream agents constrained by KL-divergence \cite{jaques2019way} and refusal training. As scalar mappings over complex dialogue, they distill multi-objective preference data into uniquely interpretable low-dimensional representations of human value. They are in essence where ``the human value rubber meets the road.'' Despite this, there is a dearth of literature analyzing the properties of reward models, largely because few have been publicly available for study. While 2023--2024 saw a proliferation of open-source language models, including Meta's Llama \cite{touvron2023llama}, Mistral AI's models \cite{jiang2023mistral}, and Google's Gemma series \cite{team2024gemma}, to date \textit{no} major industry or nonprofit lab has openly released a reward model. Only recently has this picture begun to change, with the release of \textsc{RewardBench} \cite{lambert2024rewardbench}---the first benchmark and leaderboard for reward models, spurring new activity among academic and open-source communities.
    
In this work on reward model interpretability, we seek to understand the consistency and faithfulness with which these models represent human values. Specifically, we make the following contributions: 

\vspace{-.5em}
\begin{itemize}
  \item We pioneer an exhaustive search over every single token in reward model vocabularies appended to a value-laden prompt, permitting the analysis of optimal and pessimal tokens across ten top-performing open-source reward models on \textsc{RewardBench} of varying sizes and architectures.
  \item We show that reward distributions exhibit systematic asymmetries, with greater sensitivity between tokens in high-scoring regions relative to low-scoring regions, and tokens with positive sentiment relative to negative sentiment. Changing to a negative-valence prompt inverts this latter bias, mirroring framing effects in humans.
  \item We establish alignment between the (biased) interpretation of human preferences by reward models and an independent source of ground-truth human preferences called \textsc{EloEverything}, where internet users volunteer judgments over concepts, people, and things from Wikipedia pages.
  \item We generalize our findings on similar biases and asymmetries to multi-token sequences using Greedy Coordinate Gradient optimization.
  \item Through these sequential analyses, we reveal (i) significant heterogeneity across similarly-trained reward models, invalidating their presumed fungibility, and (ii) systematic devaluation of identity group references (``homosexuals,'' ``Black people'' and ``Jews''), possibly arising as unintended biases from harmlessness training objectives.
\end{itemize}

Through exposing the idiosyncrasies of reward models as research artifacts, we aim to highlight pitfalls in their current development, provide recommendations for building more robust proxies of human value, and ultimately advance the broader goal of creating aligned and safe AI systems.

%% file: sections/ranking-tokens.tex
\section{Ranking Optimal and Pessimal Tokens} \label{sec-ranking-tokens}

\input{tables/models.tex}

Open-source reward models that convert tokens to a single scalar reward permit the somewhat radical idea of \emph{exhaustive} search. Inspired loosely by \citet{dawson2014there}, who demonstrated that exhaustive testing of floating-point math functions is feasible (arguing that ``there are only four billion floats---so test them all''), we seek a \emph{complete} ranking of rewards assigned to all possible responses to a user prompt in order to granularly and comprehensively evaluate reward model characteristics.

We used the reward-model benchmark \textsc{RewardBench} \cite{lambert2024rewardbench}, which provides an online leaderboard for top-performing reward models. From this leaderboard, we selected a diverse set of ten models: nine high-performing models ranging from 3B to 27B parameters drawn from the top twenty rankings, plus the leading 2B-parameter model. These models cover a range of developers, base architectures, and model sizes (see Table~\ref{tab:models}). 
    
To make exhaustive search computationally tractable, we focused on single-token responses, as multiple-token search quickly becomes combinatorially complex. We designed an initial prompt that specifically elicits brief responses with clear valence:\footnote{We focused our analyses on English, as reward models are predominantly trained on English data. Additional analyses across prompt variants are presented in Sec.~\ref{sec-framing-effects}.}

\begin{quote}
    \texttt{What, in one word, is the greatest thing ever?}
\end{quote}
    
We then computed the reward model score when supplying \emph{every single} token in the model's vocabulary ($N\approx256{,}000$ in the case of Gemma and $N\approx128{,}000$ in the case of Llama 3) as a response to this prompt. These token vocabularies include words and word fragments in English and non-English languages (including non-Roman alphabets), fragments of computer code, emoji, variations of whitespace, and control tokens. Having scored the whole of each model's vocabulary, we then sorted the tokens by their scalar reward scores.

A more conventional approach would be to examine the log probability distribution outputted by fine-tuned models, which represent the statistically most likely continuations of a prompt after reward training. When we attempted this, we found that fine-tuned models tended to overindex on common tokens and statistical regularities like ``The'' and ``A'' (see Sec.~\ref{sup-sec:logprobs} and Table~\ref{sup-tab:gemma-greatest}; however we also note that variations in the prompt might have produced more comparable answers to those reported here for reward models). Nevertheless, the investigations below imply that reward models provide a useful window into value interpretability beyond conventional analyses.

\input{tables/optimal-pessimal-tokens-intro}

\subsection{Qualitative Observations}
Applying this methodology to ten reward models reveals stark qualitative differences in the token rankings between models---even those from the same developer. We report optimal and pessimal tokens for two such models (\modelname{R-Gem-2B} and \modelname{R-Lla-3B}) in Table~\ref{tab:optimal-pessimal-tokens-intro-table}, and present all other models in Tables~\ref{sup-tab:raw-greatest}--\ref{sup-tab:shared-greatest}.

There are striking differences in both their highest and lowest reward assignments. At the positive extreme, \modelname{R-Gem-2B} prioritizes affective content over grammatical correctness (e.g., ranking ``miraculous'' above ``miracle''). It also prominently features the surprisingly obscure word ``sonder,'' a neologism coined in 2012 by writer John Koenig to mean ``the realization that each random passerby is living a life as vivid and complex as your own'' \cite{koenig2021dictionary}. The \modelname{R-Lla-3B} model from the same developer instead puts ``freedom'' ahead of ``love,'' and on the whole the high rankings of ``freedom,'' ``opportunity,'' ``discovery,'' and ``experience'' paint a more individualistic, active picture of human value than \modelname{R-Gem-2B}'s more interdependent, affective words like ``love,'' ``wonder,'' and ``hope.''

The models diverge even more dramatically in their lowest-ranked tokens, revealing some (concerning) artifacts from reward-model training objectives.  The lower ranks of the \modelname{R-Gem-2B} model are tied to human harm and suffering like ``rape,'' ``Hitler,'' and ``murderers,'' as well as slurs and profanities, suggesting a strong influence of a harmlessness objective in training. In contrast, \modelname{R-Lla-3B}'s lowest ranks are predominantly occupied by malformed code tokens and programming artifacts like ``.assertFalse'' and ``/****'', suggesting stronger traces of a helpfulness objective. Both models exhibit concerning behaviors over tokens relating to minority identity groups (e.g., ``blacks'' or ``homosexuals'').  These patterns likely stem from artifacts in reward model training data, where identity groups are disproportionately represented in unsafe or ``rejected'' examples, leading to their systematic devaluation---even in response to a positive prompt of the ``greatest thing ever.'' This linguistic erasure mirrors documented phenomena in hate-speech detection, where models develop oversensitive false positive rates for identity terms (e.g., ``Muslim,'' ``mosque'') or reclaimed slurs due to their overrepresentation in negative training contexts and underrepresentation in neutral or positive ones \citep{dixon2018measuring, park-etal-2018-reducing, sap2019risk}.

\subsection{Quantitative Analysis}

\begin{figure}[t!]
    \centering
    {\includegraphics[width=\columnwidth]{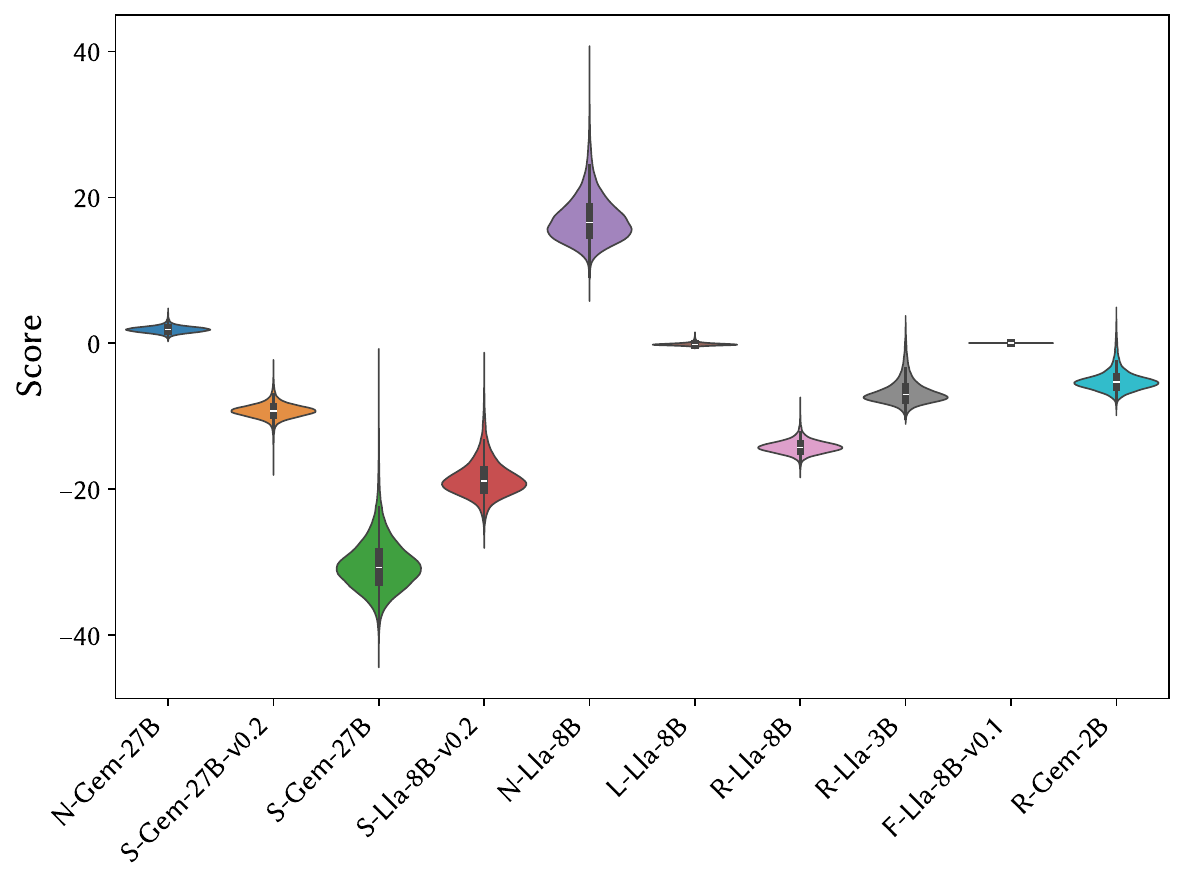}}
    \caption{Violin plot of exhaustive score distributions to the ``greatest thing'' prompt. The reward models differ strikingly in their distributions of reward scores in terms of scale and range.}
    \label{fig:reward-model-distributions}
\end{figure}

\noindent Quantitatively, we note that, despite differences of the scale of scores across models (Fig.~\ref{fig:reward-model-distributions}), all score distributions exhibit a positive skew (Table~\ref{tab:moments}). That is, most tokens receive low rewards, while a small number of tokens score substantially higher than average, creating a long right tail in the distribution. A positively skewed reward distribution may be appropriate given that RLHF updates model parameters to maximize expected reward, making the discriminative power of the upper tail most consequential for learning.

\begin{figure*}[t!]
    \centering
    \begin{subfigure}[t]{\textwidth}
        \begin{tikzpicture}[baseline=(image.north)]
            \node[anchor=south west,inner sep=0] (image) at (0,0) {\includegraphics[height=5.8cm]{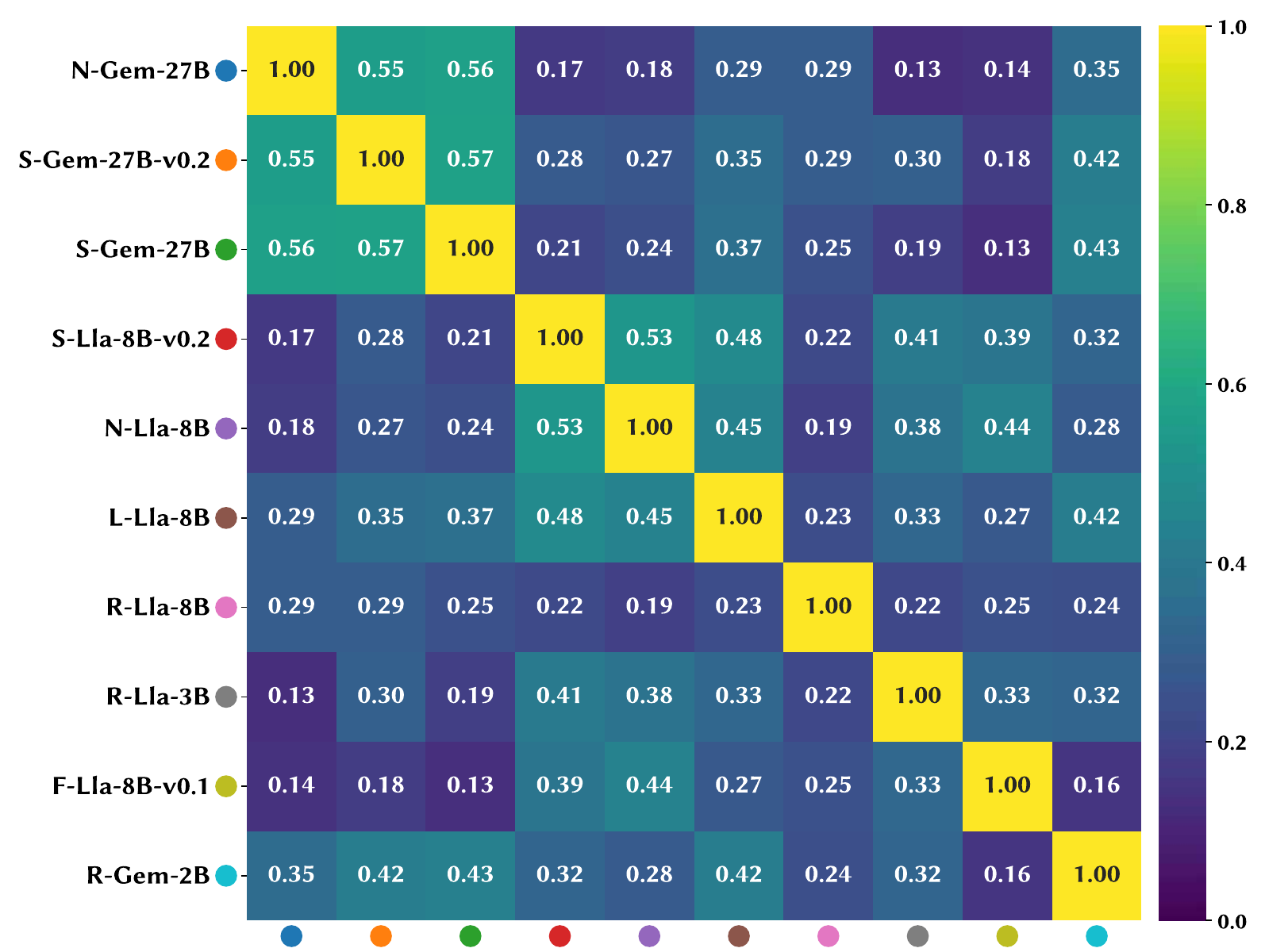}};
            \begin{scope}[x={(image.south east)},y={(image.north west)}]
                \node[font={\sffamily\Large}] at (-0.02,0.95) {A};
            \end{scope}
        \end{tikzpicture}
        \hfill
        \begin{tikzpicture}[baseline=(image.north)]
            \node[anchor=south west,inner sep=0] (image) at (0,0) {\includegraphics[height=5.4cm]{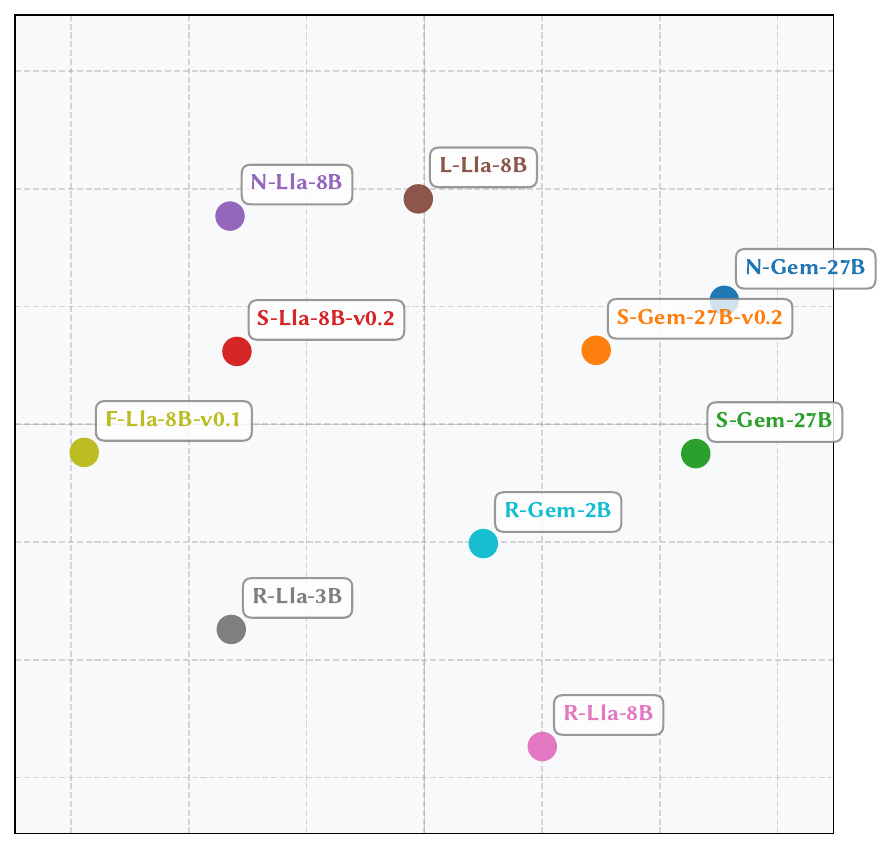}};
            \begin{scope}[x={(image.south east)},y={(image.north west)}]
                \node[font={\sffamily\Large}] at (-0.05,0.95) {B};
            \end{scope}
        \end{tikzpicture}
        \hfill
        \begin{tikzpicture}[baseline=(image.north)]
            \node[anchor=south west,xshift=0.6cm] (image) at (0,0) {\includegraphics[height=5.6cm]{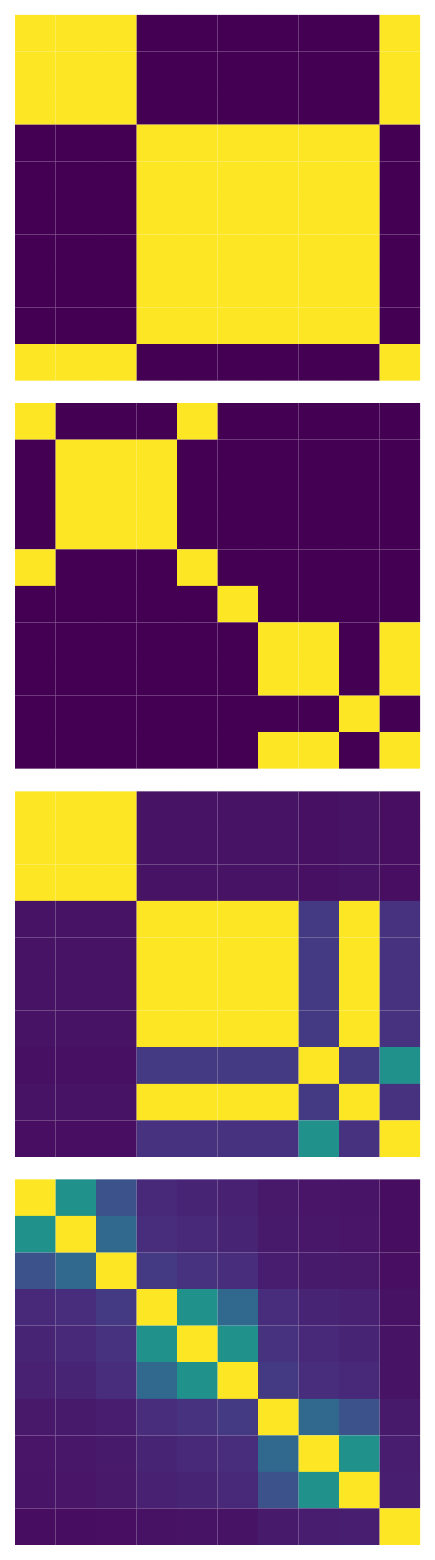}};
            \begin{scope}[x={(image.south east)},y={(image.north west)}]
                \node[font={\sffamily\Large}] at (-0.05,0.95) {C};
            \end{scope}
        \end{tikzpicture}
    \end{subfigure}

    \caption{(A) Heatmap depicting the pairwise Kendall's $\tau$ correlations between the reward models for scored responses to the prompt ``What, in one word, is the greatest thing ever?''. (B) Visualization of the degree of similarity between reward models using multidimensional scaling (MDS) of the Kendall's $\tau$ distance measure. (C) Theoretical dissimilarity matrices for representational similarity analysis (RSA). The four dissimilarity matrices encode, respectively, base model $[\mathrm{base}_i=\mathrm{base}_j]$; developer $[\mathrm{dev}_i=\mathrm{dev}_j]$; parameter count $(1+|\mathrm{params}_i-\mathrm{params}_j|)^{-1}$; and RewardBench ranking $(1+|\mathrm{rank}_i-\mathrm{rank}_j|)^{-1}$.}
    \label{fig:model_similarity}
\end{figure*}

We assessed the consistency of token rankings across models using an ordinal correlation measure, Kendall's $\tau$, Fig.~\ref{fig:model_similarity}A (results are consistent across choice of correlation metric, see Fig.~\ref{sup-fig:correlation-heatmaps-spearman-rbo-greatest}). Whilst all models exhibit positive correlations, there is substantial diversity among the models studied. We explored this diversity using multidimensional scaling (MDS)---a visualization technique that aims to faithfully represent the degree of similarity between data points (here, reward models) in lower dimensionality (here, 2D). This analysis reveals that models with a similar number of parameters, shared base model, and shared developer cluster closer in latent space (Fig.~\ref{fig:model_similarity}B).

\input{tables/reward_distribution_moments}

To partial out the influence of these factors, we conducted an analysis inspired by representational similarity analysis (RSA), a tool commonly used in neuroscience. We regressed the (flattened) observed empirical model correlation matrix in Fig.~\ref{fig:model_similarity}A on theoretical model similarity matrices based on the three factors of interest (base model, developer, number of parameters) and the rank of the model on the \textsc{RewardBench} leaderboard (Fig.~\ref{fig:model_similarity}C). Each of these factors is, on its own, significantly associated with the empirical pattern of correlations between models (simple linear regression, all $p$ < .001). However, when combining the four factors together in a competitive multiple regression, the variance predicted by base model, developer, and the number of parameters appears to be almost entirely soaked up by the ranking of the model on \textsc{RewardBench} (\textsc{RewardBench} rank $p$ < .0001, base model $p$ < .10, all other $p$ > .10). Running a stepwise regression (factor knock-in and knock-out) confirms that the regression that best explains the observed data features the base model and \textsc{RewardBench} ranking theoretical matrices (base model $\beta$ = 0.05, \textsc{RewardBench} ranking $\beta$ = 0.69, $R^2$ = .80). It is perhaps unsurprising that model ranking on \textsc{RewardBench} can capture patterns of reward model similarity since it measures how well the models are all aligned against the same external objective (i.e.,\ the \textsc{RewardBench} benchmark and its composite evaluation datasets).
Interestingly, our results suggest that the choice of base model drives differences in token rankings above and beyond alignment to \textsc{RewardBench}. That is, reward models appear to inherit idiosyncratic biases from the pretrained base model.

%% file: tables/models.tex
\begin{table*}
\caption{Open-source reward models studied. The table includes both their full names and the shortened identifiers (Model IDs) used throughout the rest of this paper. Ranks are from the RewardBench Leaderboard as of January 14, 2025.}
\small
\centering
    \begin{tabular}{rllllr}
    \toprule
    RewardBench Rank & \hspace{0.6em} Model ID & Developer & Model Name & Base Model & Parameters (B) \\
    \midrule
    2 & \modelname{N-Gem-27B} & nicolinho & QRM-Gemma-2-27B\cite{dorka2024quantile} & Gemma 2\cite{team2024gemma2} & 27 \\
    3 & \modelname{S-Gem-27B-v0.2} & Skywork & Skywork-Reward-Gemma-2-27B-v0.2\cite{liu2024skywork} & Gemma 2 & 27 \\
    5 & \modelname{S-Gem-27B} & Skywork & Skywork-Reward-Gemma-2-27B\cite{liu2024skywork} & Gemma 2 & 27 \\
    10 & \modelname{S-Lla-8B-v0.2} & Skywork & Skywork-Reward-Llama-3.1-8B-v0.2\cite{liu2024skywork} & Llama 3.1\cite{dubey2024llama} & 8 \\
    11 & \modelname{N-Lla-8B} & nicolinho & QRM-Llama3.1-8B\cite{dorka2024quantile} & Llama 3.1 & 8 \\
    12 & \modelname{L-Lla-8B} & LxzGordon & URM-LLaMa-3.1-8B\cite{lou2024uncertainty} & Llama 3.1 & 8 \\
    17 & \modelname{R-Lla-8B} & Ray2333 & GRM-Llama3-8B-rewardmodel-ft\cite{yang2024regularizing} & Llama 3 & 8 \\
    19 & \modelname{R-Lla-3B} & Ray2333 & GRM-Llama3.2-3B-rewardmodel-ft\cite{yang2024regularizing} & Llama 3.2 & 3 \\
    20 & \modelname{F-Lla-8B-v0.1} & RLHFlow & ArmoRM-Llama3-8B-v0.1\cite{wang2024interpretable} & Llama 3 & 8 \\
    31 & \modelname{R-Gem-2B} & Ray2333 & GRM-Gemma2-2B-rewardmodel-ft\cite{yang2024regularizing} & Gemma 2 & 2 \\
    \bottomrule
    \end{tabular}
\label{tab:models}
\end{table*}

%% file: tables/optimal-pessimal-tokens-intro.tex
\begin{table*}
\footnotesize
\centering
\caption{Optimal and pessimal tokens for the prompt ``What, in one word, is the greatest thing ever?''. Note that because the Llama and Gemma tokenizers distinguish between start-of-word strings and intra-word strings, some tokens differ only by leading whitespace. \textcolor{red}{CONTENT WARNING: We present tokens in their raw form (including slurs) to enable transparent attribution of model tokens, while acknowledging their offensive, troubling and harmful nature.}}
\label{tab:optimal-pessimal-tokens-intro-table}
\begin{minipage}{0.42\textwidth}
\centering
  \caption*{R-Gem-2B}\vspace{-1em}
  \label{tab:top-bottom-greatest--gemma-2-9b-}
  \begin{tabular}{rll}
\toprule
Token ID & Decoded & Score \\
\midrule
27534 & \texttt{ LOVE} & 4.594 \\
61792 & \texttt{LOVE} & 4.562 \\
218136 & \texttt{ felicity} & 4.469 \\
2182 & \texttt{ love} & 4.344 \\
12870 & \texttt{love} & 4.312 \\
7377 & \texttt{ Love} & 4.281 \\
8703 & \texttt{Love} & 4.281 \\
227570 & \texttt{sonder} & 4.219 \\
143735 & \texttt{ sonder} & 4.219 \\
27539 & \texttt{ Wonder} & 4.188 \\
34183 & \texttt{Wonder} & 4.188 \\
174540 & \texttt{HOPE} & 4.156 \\
115221 & \texttt{ HOPE} & 4.125 \\
5144 & \texttt{ wonder} & 4.094 \\
53798 & \texttt{wonder} & 4.094 \\
167954 & \texttt{ WONDER} & 4.031 \\
50999 & \texttt{ bliss} & 3.922 \\
207783 & \texttt{bliss} & 3.922 \\
65646 & \texttt{JOY} & 3.922 \\
135936 & \texttt{ JOY} & 3.922 \\
89399 & \texttt{ miraculous} & 3.875 \\
40241 & \texttt{ miracle} & 3.859 \\
... & \texttt{...} & ... \\
61001 & \texttt{ blacks} & -9.250 \\
218552 & \texttt{ pathologist} & -9.250 \\
97070 & \texttt{ killers} & -9.312 \\
167921 & \texttt{ prostitutes} & -9.312 \\
222988 & \texttt{ massacres} & -9.312 \\
106863 & \texttt{ FUCKING} & -9.312 \\
213624 & \texttt{ rapist} & -9.312 \\
127732 & \texttt{ ransomware} & -9.375 \\
204573 & \texttt{ retards} & -9.438 \\
195353 & \texttt{ nazis} & -9.438 \\
137696 & \texttt{ murdering} & -9.438 \\
37678 & \texttt{ Hitler} & -9.500 \\
230672 & \texttt{Rape} & -9.500 \\
134768 & \texttt{ Rape} & -9.500 \\
231158 & \texttt{ faggot} & -9.500 \\
144817 & \texttt{ murderous} & -9.500 \\
152471 & \texttt{ murderers} & -9.500 \\
39688 & \texttt{ rape} & -9.562 \\
144068 & \texttt{Hitler} & -9.562 \\
186353 & \texttt{rape} & -9.625 \\
158058 & \texttt{ negroes} & -9.625 \\
201371 & \texttt{ raping} & -9.625 \\
\bottomrule
  \end{tabular}
\end{minipage}
\hfill
\begin{minipage}{0.52\textwidth}
\centering
    \caption*{R-Lla-3B}\vspace{-1em}
    \label{tab:top-bottom-greatest--R-Lla-3B}
    \begin{tabular}{rll}
\toprule
Token ID & Decoded & Score \\
\midrule
11542 & \texttt{ freedom} & 3.359 \\
86872 & \texttt{Freedom} & 3.266 \\
25320 & \texttt{ Freedom} & 3.266 \\
40835 & \texttt{ LOVE} & 3.250 \\
61094 & \texttt{ LIFE} & 3.203 \\
83900 & \texttt{.life} & 3.000 \\
24966 & \texttt{ CONNECTION} & 2.969 \\
28899 & \texttt{ imagination} & 2.844 \\
10919 & \texttt{ Love} & 2.672 \\
29351 & \texttt{Love} & 2.672 \\
48379 & \texttt{ Opportunity} & 2.641 \\
57184 & \texttt{ UNITY} & 2.438 \\
57273 & \texttt{UNITY} & 2.438 \\
6776 & \texttt{ opportunity} & 2.406 \\
32883 & \texttt{ loyalty} & 2.359 \\
3021 & \texttt{ love} & 2.281 \\
31153 & \texttt{love} & 2.281 \\
39193 & \texttt{ Discovery} & 2.266 \\
68500 & \texttt{Discovery} & 2.266 \\
35215 & \texttt{Experience} & 2.234 \\
21460 & \texttt{ Experience} & 2.234 \\
85743 & \texttt{Peace} & 2.156 \\
... & \texttt{...} & ... \\
87546 & \texttt{ raping} & -10.688 \\
86395 & \texttt{.FindGameObjectWithTag} & -10.688 \\
38853 & \texttt{\$\{} & -10.688 \\
18350 & \texttt{(con} & -10.750 \\
27817 & \texttt{\_headers} & -10.750 \\
58467 & \texttt{.insertBefore} & -10.750 \\
6019 & \texttt{(st} & -10.750 \\
29372 & \texttt{(cfg} & -10.750 \\
5747 & \texttt{.setText} & -10.750 \\
27701 & \texttt{.startsWith} & -10.750 \\
26342 & \texttt{/******************...} & -10.812 \\
97615 & \texttt{ \#\#\#\#\#\#\#\#\#\#\#\#\#\#\#\#\#\#...} & -10.812 \\
85399 & \texttt{\#\#\#\#\#\#\#\#\#\#\#\#\#\#\#\#\#\#\#...} & -10.812 \\
76897 & \texttt{\_checks} & -10.875 \\
58352 & \texttt{("[\%} & -10.875 \\
74061 & \texttt{/******************...} & -10.938 \\
42864 & \texttt{ homosexual} & -10.938 \\
6294 & \texttt{(struct} & -10.938 \\
27249 & \texttt{.startswith} & -11.000 \\
94380 & \texttt{ jihadists} & -11.062 \\
97223 & \texttt{ homosexuals} & -11.312 \\
37289 & \texttt{.assertFalse} & -11.438 \\
\bottomrule
\end{tabular}
\end{minipage}
\end{table*}

%% file: tables/reward_distribution_moments.tex
\begin{table}[t!]
\begin{tabular}{lrrr}
\toprule\addlinespace
\hspace{0.6em} Model	        &	Mean	    &	Variance	&	Skewness \\
\addlinespace\midrule\addlinespace
\modelname{N-Gem-27B}	    &	1.876	    &	0.193    	&	0.437 \\
\modelname{S-Gem-27B-v0.2}  &	-9.274	    &	1.017    	&	0.071 \\
\modelname{S-Gem-27B}	    &	-30.422  	&	11.991   	&	0.878 \\
\modelname{S-Lla-8B-v0.2}	&	-18.699  	&	5.716    	&	1.117 \\
\modelname{N-Lla-8B}	    &	16.613    	&	9.558   	&	1.133 \\
\modelname{L-Lla-8B}	    &	-0.137   	&	0.034    	&	1.763 \\
\modelname{R-Lla-8B}	    &	-14.239   	&	0.781    	&	0.504 \\
\modelname{R-Lla-3B}	    &	-6.777   	&	2.597    	&	1.672 \\
\modelname{F-Lla-8B-v0.1}	&	0.031   	&	<0.001    	&	1.055 \\
\modelname{R-Gem-2B}	    &	-5.279   	&	1.957    	&	1.457 \\
\addlinespace\bottomrule
\end{tabular}
\caption{First three moments of reward distribution across all shared tokens. All reward models exhibit varying degrees of positive skew.}
\label{tab:moments}
\end{table}


%% file: sections/framing.tex
\vspace{-.5\baselineskip}
\begin{figure*}[!t]
    \includegraphics[width=\textwidth]{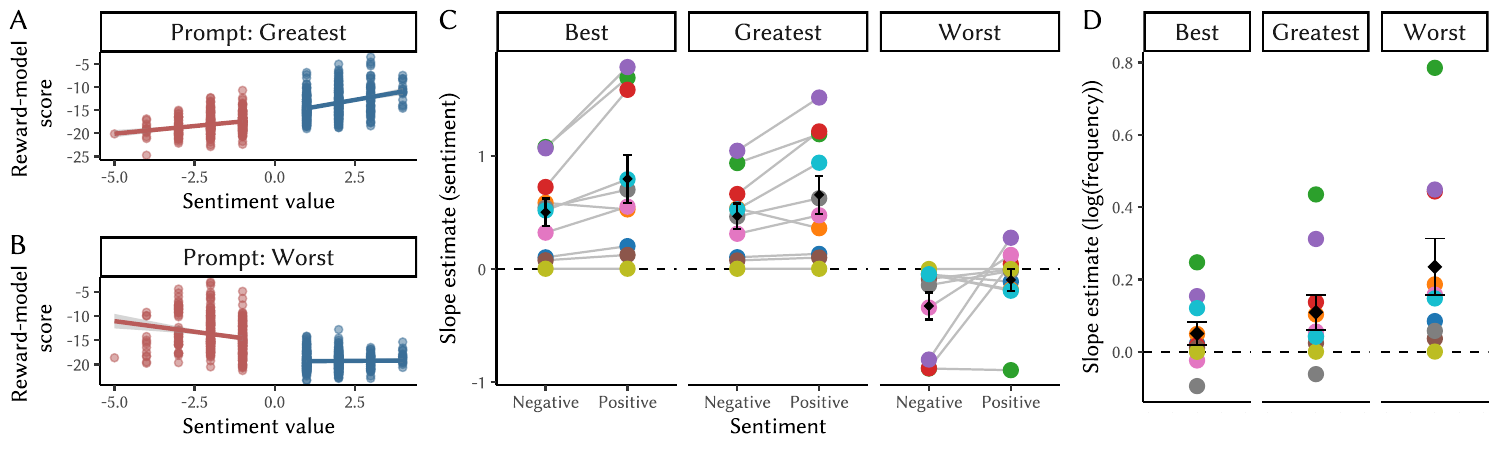}
    \caption{(A) Correlation plot between token sentiment value according to the AFINN-111 lexicon and the scores from the \modelname{S-Lla-8B-v0.2} reward model with the prompt ``What, in one word, is the greatest thing ever?'' (B) As previous, but for prompt ``What, in one word, is the worst thing ever?'' (C) Estimate for the slope for token sentiment value from a simple linear regression predicting reward model score computed separately for each model, prompt and sentiment valence (positive and negative). Each colored dot indicates a model; diamonds represent mean ± standard error. Slope estimates are, on average, higher for positive sentiment. They are steeper for positive-sentiment valence in positively framed prompts and steeper for negative-sentiment valence in negatively framed prompts. (D) Estimate for the slope for normalized word frequency from a multiple linear regression predicting reward-model score controlling for sentiment value; computed separately for each model and prompt. Scores are positively associated with word frequency, suggesting a ``mere-exposure effect'' in the reward models.}
    \label{fig:sentiment-slopes}
\end{figure*}

\section{Framing Effects}
\label{sec-framing-effects}

\subsection{Sentiment Analysis}
To further explore explanations for token rankings, we investigated the relationship between sentiment, or the emotional value of a token, and its reward model score. We quantified emotional value using data from two validated linguistic corpora widely used within the field of psychology and developed by human experts: \textsc{Bing} \cite{liu2022sentiment} and \textsc{AFINN-111} \cite{IMM2011-06010}. \textsc{Bing} codes words as ``positive'' or ``negative''; \textsc{AFINN-111} indexes a score ranging from $-5$ to $5$ for the sentiment value. Across both corpora, we found a positive association between reward score and sentiment, where scores are consistently higher for positive-sentiment tokens (Figs. \ref{fig:sentiment-slopes}A, \ref{sup-fig:sentiment-bing}--\ref{sup-fig:sentiment-afinn}). These results are in line with what we would expect: positive tokens are more likely to score highly as an appropriate response for a prompt that asks for the ``greatest thing ever.'' In line with the skewness of the score distribution in Sec.~\ref{sec-ranking-tokens}, we found that (i) scores for positive-sentiment tokens are more spread out than scores for negative-sentiment tokens, and (ii) the slope for the relationship between sentiment and score is significantly steeper for positive than negative tokens (Fig.~\ref{fig:sentiment-slopes}A; $\beta_\mathrm{pos}$ > $\beta_\mathrm{neg}$: $t$(9) = 2.6, $p$ < 0.05). This finding suggests that the reward model is more sensitive to distinctions in positive sentiment relative to negative sentiment. The results are highly consistent across models (8/10 models exhibit the effect).
\vspace{-1em}
\subsection{Prompt Framing}
To explore whether the differential sensitivity of the model is driven by the specifics of the prompt or generalizes across queries, we extended our analyses to two more prompts: a positive variant (``What, in one word, is the best thing ever?'') and a negative contrast (``What, in one word, is the worst thing ever?''). Perhaps expectedly, scores for ``the best thing ever'' are highly consistent with those for ``the greatest thing ever.'' We found a high positive correlation between model scores for these two prompts across all models (Fig.~\ref{fig:best-vs-greatest}). We replicated (i) the positive skewness of the distribution of scores and (ii) the differential sensitivity to positive over negative-sentiment tokens (Fig.~\ref{fig:sentiment-slopes}C; 9/10 models exhibit the effect). 

The pattern is different for ``the worst thing ever.'' Here, the distribution of scores remains skewed toward higher-scoring tokens, however, it is the \textit{negative-sentiment} tokens that receive higher rewards for this prompt. Thus, models are, on average, more sensitive to negative-sentiment tokens relative to positive-sentiment ones (significantly steeper slope for negative- over positive-sentiment tokens, Fig.~\ref{fig:sentiment-slopes}B--C; 5/10 models exhibit the effect). Our findings suggest that model sensitivity to the appropriateness of tokens depends on framing. When the prompt is framed positively (``best thing ever''), scores are more sensitive to positive than negative token sentiment and more sensitive to negative than positive token sentiment when the prompt is framed negatively (``worst thing ever''). This result is consistent with human behavior. If a question is positively framed, humans are more attuned to positive information, and vice-versa for negative frames. Consider a scenario where you need to pick between two vacation destinations: an exciting option with many positive features (dream destination, beautiful nature) and just as many drawbacks (expensive, long travel) versus a safer option with fewer positive and negative stand-out features (e.g.,\ a local getaway). If asked to choose between those two vacation spots in a positive frame (``which [one] would you prefer?''), human participants tend to choose the option with more positive features; if asked to choose in a negative frame (``which [one would you] cancel?''), they pick the option with more negative features, even though it is in fact the same option \cite{shafir1993reason}.

 The effect of framing on sensitivity has important implications. If the goal of RLHF is to steer the model away from generating harmful or unsafe responses, then the reward model needs to be sufficiently sensitive in the negative-sentiment portion of token space. Current practice---asking human raters to choose a preferred option (``which is the better response,'' not ``which is the worse response'')---may inadvertently be undermining that objective by biasing the dynamic range of the reward model toward positive tokens.

  \begin{figure*}[t!]
    \centering
    \includegraphics[width=.8\textwidth]{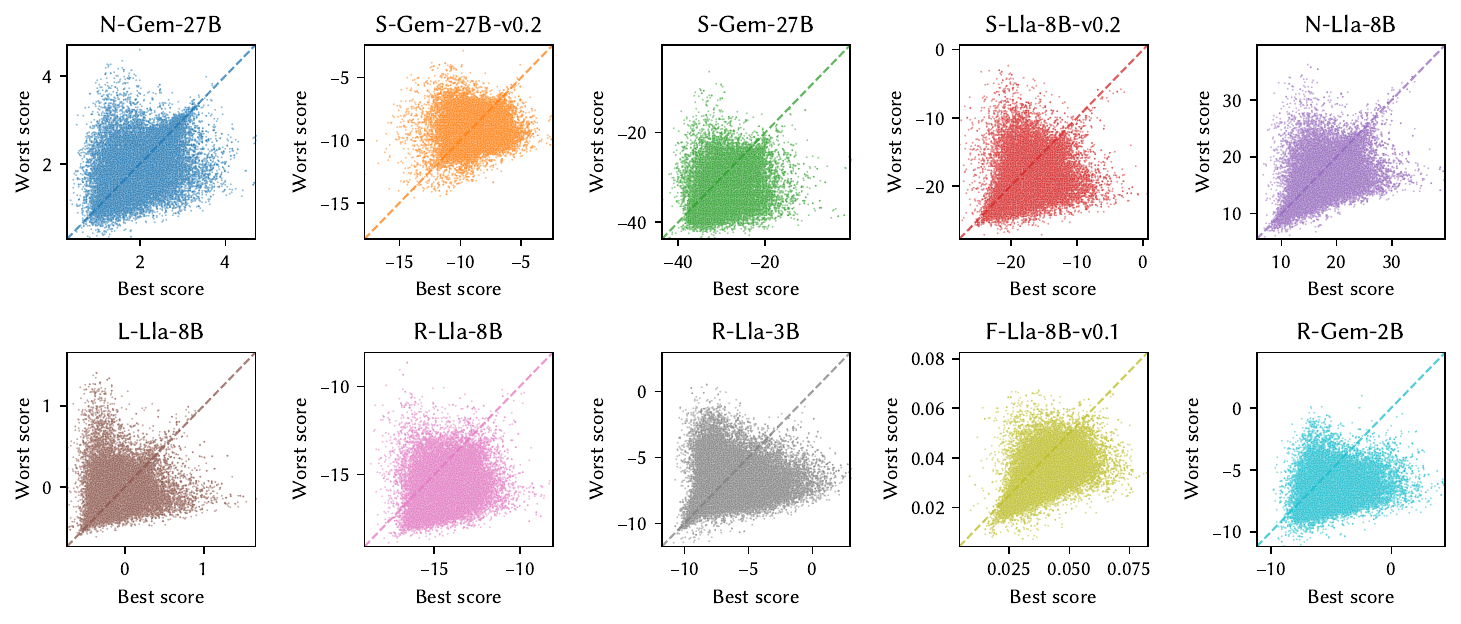}
    \vspace{-0.5cm}
    \caption{Juxtaposing exhaustive scores for the ``best thing'' prompt against the ``worst thing'' prompt reveals not just a simple negative correlation, but also an orthogonal dimension representing tokens that are bad or good responses to \emph{both} frames.}
    \label{fig:best-vs-worst}
    \vspace{-1em}
  \end{figure*}

  Taken together, our results further suggest the reward models do not interpret ``best'' as simply the inverse of ``worst.'' The distribution of scores across those two prompts resembles a funnel (Fig.~\ref{fig:best-vs-worst}) where many tokens are bad responses to both prompts (bottom left) and some tokens are good responses to one prompt but not the other (top left and bottom right). We also see a thin tail of tokens that are highly-scored responses for \emph{both} prompts (top right;  ``sonder'' features in this category, along with non-committal answers like ``depends'' and refusals like ``impossible''). In the appendix, we include tables that index the $\mathrm{best} + \mathrm{worst}$ (tokens that score similarly on both prompts) and $\mathrm{best} - \mathrm{worst}$ axes (tokens that score highly on one but not the other prompt); see Tables~\ref{sup-tab:shared-best-plus-worst}--\ref{sup-tab:shared-best-minus-worst}.
  
\vspace{-1em}
\subsection{Frequency Bias}
  Are the scores that reward models assign to different tokens biased by how frequently the word appears in the English language? To assess this, we used data from Word Frequencies in Written and Spoken English \cite{leech2014word} and regressed log-transformed word frequency on reward-model scores. Higher word frequency is associated with higher reward-model score across prompts for the majority of models (positive slope estimates with $p$ < .05 in 10/10 models for ``best,'' 8/10 for ``greatest,'' and 7/10 for ``worst''). This is reminiscent of the ``mere-exposure effect'' in humans, where the more someone is exposed to a stimulus, the more they like it \cite{zajonc1968attitudinal}. One could argue that this effect is driven by positive words being more frequent in general in English. To account for this, we controlled for the sentiment value of the tokens. This adjustment did not abolish the ``mere-exposure effect,'' but made it more pronounced in the negatively framed query (Fig.~\ref{fig:sentiment-slopes}D, positive slope estimates with $p$ < .05 in 2/10 models for ``best,'' 5/10 for ``greatest'' and 8/10 for ``worst'').
  
  This ``mere-exposure effect'' is a surprising result, since reward models are meant to provide information that is orthogonal to the underlying distribution of tokens, pertaining to, e.g., helpfulness and harmlessness. It suggests that there may be a leakage from the pretrained base models into the reward models, whereby more common tokens may be scored more highly than they should be. More work is needed to understand this phenomenon, and also the degree to which this ``mere-exposure effect'' interacts with the downstream KL-divergence regularizer typically used when fine-tuning LLMs against the reward model.

%% file: sections/comparison-to-humans.tex
\section{Alignment with \textsc{EloEverything}}
\begin{figure*}
    \centering\includegraphics[width=\linewidth]{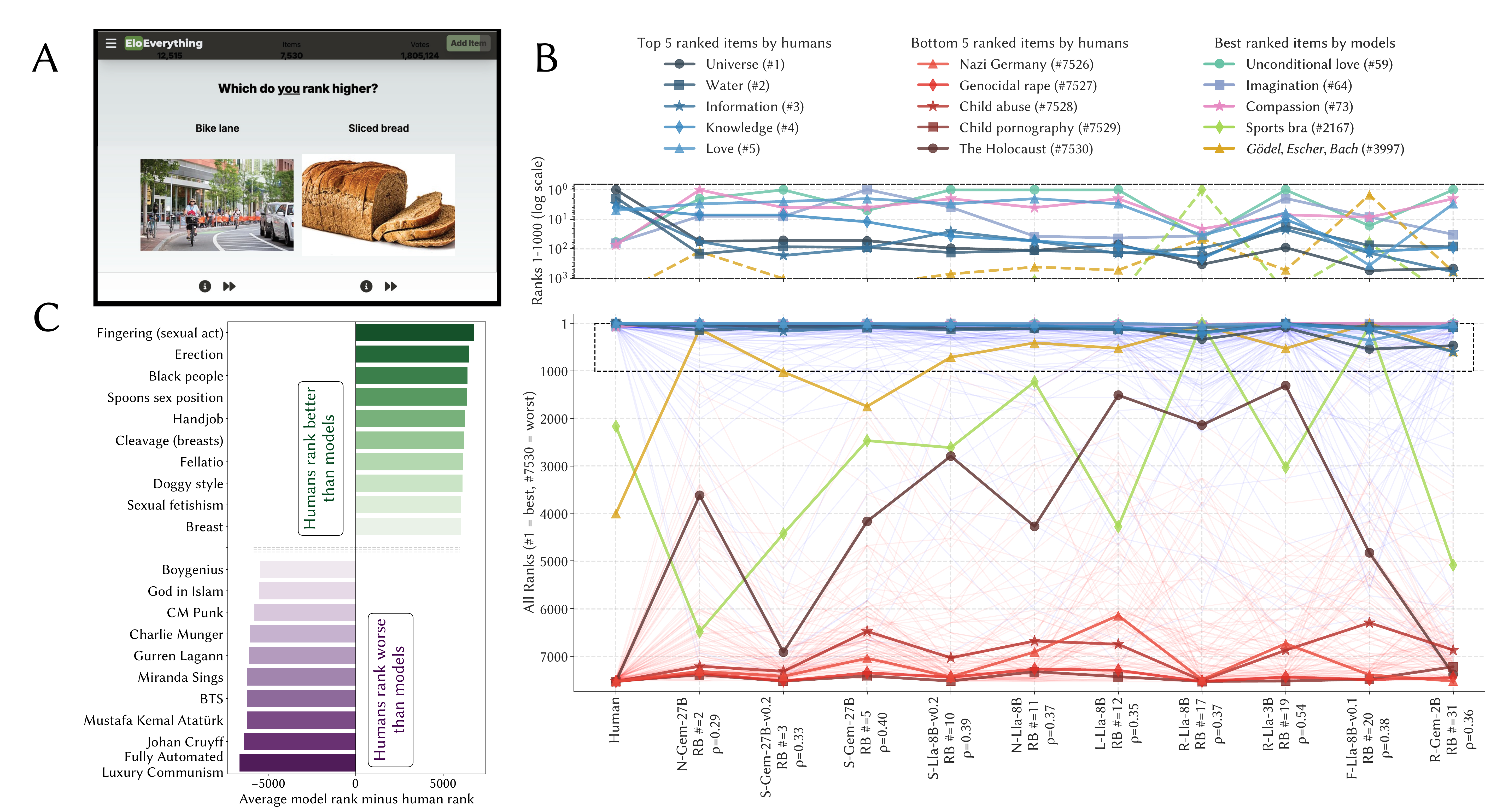}
    \caption{(A) The \textsc{EloEverything} ranking interface where users make pairwise preference judgments between items (e.g., ``Bike lane'' vs ``Sliced bread''). (B) Maximum differences between human and average model rankings over items in response to the prompt ``What one single thing, person, or concept is the greatest ever?'', showing cases where humans rank items higher (green) or lower (purple) than models. (C) Rank trajectory plot showing how human and model ranks differ. We plot (i) the top 5 items in the human rank (blue color scale with human ranks shown in legend parentheses as $\#$n), (ii) the bottom 5 items in the human rank (red color scale), and (iii) unique items ranked \#1 by models. Specifically, ``Unconditional love'' is \#1 for 5 models; ``Compassion'' is \#1 for \modelname{N-Gem-27B}; ``Imagination'' for \modelname{S-Gem-27B}; ``Sports bra'' for \modelname{R-Lla-8B}; and ``\textit{Gödel, Escher, Bach}'' for \modelname{F-Lla-8B-v0.1}. Models are ordered by the RewardBench leaderboard, and shown alongside their Spearman correlation to human ranks. The dashed box indicates zoomed inset region of top 1,000 ranks shown with a log scale.}
    \label{fig-ee-everything}
\end{figure*}

Thus far we have identified internal inconsistencies within and across reward models. However, reward models are intended to proxy human value judgments. Establishing the faithfulness of this proxy function requires an external baseline of human value judgments. We sourced this external human preference data from \textsc{EloEverything},\footnote{\url{https://eloeverything.co/}.} a crowdsourcing platform that implements pairwise preference learning over things, people, and concepts uploaded from Wikipedia. On \textsc{EloEverything}, internet users are presented with pairs of Wikipedia-derived entities (accompanied by images) and volunteer their judgments in response to the prompt: ``Which do \uline{you} rank higher?'' (see Fig.~\ref{fig-ee-everything}A), with options to request additional context or skip. The platform aggregates these pairwise comparisons using the Elo rating system, which was originally developed for chess rankings \cite{elo1967proposed} and has since been widely adopted to evaluate LLMs \cite{bai2022traininghelpfulharmlessassistant, boubdir2023elouncoveredrobustnessbest, chiang2024chatbotarenaopenplatform}. We collected all data from the \textsc{EloEverything} website, resulting in a dataset that comprises $N_{\mathrm{users}}$ = 12,515 users who evaluate $N_{\mathrm{items}}$ = 7,530 items across $N_{\mathrm{pairings}}$ = 1,805,124 total pairwise comparisons. Although the dataset is highly imbalanced by the ratings each item receives ($\mu$ = 479.4 pairings/item, $\sigma$ = 464.4)\footnote{This is because users can upload new items at any point. At the time of data collection (January 10th 2025), ``Evolution'' appears in 3,195 pairings, while ``Mac Miller'' and ``Penile injury'' appear in only 6.} and likely imbalanced across non-representative users (user-level data is not available), it still serves as a valuable independent baseline.

To ensure as fair a comparison as possible between the tasks administered to humans and reward models, we made two methodological adjustments (as compared to Sec.~\ref{sec-ranking-tokens}). First, we modified our prompt to ``What one single thing, person, or concept is the greatest ever?'' to better align with the human task and accommodate the multi-word concepts that appear in \textsc{EloEverything}, such as ``Sliced bread,'' ``Female body shape,'' ``Freedom of the press'' or even ``Beliefs and practices of the Church of Jesus Christ of Latter-day Saints'' ($\mu$ = 2.1 words/item, $\sigma$ = 1.3).\footnote{To test generalizability across prompts, we also repeated analysis in this section for our original prompt (``What, in one word, is the greatest thing ever?'') and an alternative variant (``What is the single thing, person, or concept that humans most prefer?''). We present results in Figs.~\ref{fig:within-prompt-xmodel}--\ref{fig:ee-kendall-elo-everything}.} Second, rather than exhaustively evaluating the entire model vocabulary, we restricted this analysis to the set of \textsc{EloEverything} items ($N_{\mathrm{items}}=7{,}350$) to ensure a common human--model comparison set. We used the same set of models as in Table~\ref{tab:models}, and normalized both reward model scores and human Elo ratings to rankings, using average rank for ties. 

\subsection{Heterogeneity and Asymmetry in Human-Model and Model-Model Alignment} Our analysis reveals several notable patterns in the relationship between the ground truth preferences of \textsc{EloEverything} raters and human preferences as interpreted by reward models. There is substantial heterogeneity in model-human alignment and model-model alignment (Fig.~\ref{fig:within-prompt-xmodel}). The mean Kendall's $\tau$ correlation between human and model rankings is 0.29 ($\sigma$ = 0.06), with coefficients ranging from 0.22 to 0.39 across models, indicating only moderate rank agreement. Divergence of reward models from human rankings might be expected---given that \textsc{EloEverything} users benefit from additional context like images and can personalize their ratings with ``Which do \uline{you} rank higher''---but the observed heterogeneity extends to model-model comparisons. The mean Kendall's $\tau$ correlation between pairs of different models is 0.55. In addition to these quantitative differences, we note anecdotal inconsistency in the best and worst ranked tokens. For example, while ``Unconditional love'' is ranked best by 5 models, ``Sports bra'' is top for \modelname{R-Lla-8B} and ``\textit{Gödel, Escher, Bach}'' for \modelname{F-Lla-8B-v0.1}. As demonstrated in (Fig.~\ref{fig-ee-everything}C), ranks display substantial movement across models, challenging the assumption that reward models trained under a similar objective can be used interchangeably.

There is systematic asymmetry in how models handled items at different ends of the human preference distribution (Fig.~\ref{fig-ee-everything}C). Reward models show stronger agreement with human rankings for highly-rated items ($\tau_{\mathrm{top100}}$ = 0.19) compared to low-rated items ($\tau_{\mathrm{bottom100}}$ = 0.08). This asymmetry (and heterogeneity) is evident in  Fig.~\ref{fig-ee-everything}C, where ``The Holocaust'' (ranked \textit{worst} by humans) is ranked substantially higher by most models, even those performing well on \textsc{RewardBench}. This corroborates our findings in Sec.~\ref{sec-ranking-tokens}, where models are more sensitive to high- over low-scoring tokens (the positive skew of the score distribution).

\subsection{Analyzing Discrepancies between Value as Perceived by Humans and Models}

Relative to human rankings, reward models systematically undervalue concepts related to nature and life, e.g., ``Universe'' (human rank $\#_{H}=1$, mean rank across models $\#_{\bar{M}}=320$), ``Gravity'' ($\#_{H}=7$, $\#_{\bar{M}}=320$), ``Breathing'' ($\#_{H}=16.5$, $\#_{\bar{M}}=321$); and technological concepts, e.g., ``Technology'' ($\#_{H}=47.5$, $\#_{\bar{M}}=633.5$), ``Electronics'' ($\#_{H}=78$, $\#_{\bar{M}}=3966$), ``Computer'' ($\#_{H}=95.5$, $\#_{\bar{M}}=1352.5$). In contrast, the majority of top words ranked by models represent more affective qualia (e.g.,\ ``Unconditional love,'' ``Imagination,'' ``Hope,'' or ``Happiness''). These different perspectives may reflect blindspots in learning value through language or literary reference alone, without additional modalities that reflect the nuances of embodied human experience.

The most striking valuation differences (see Fig.~\ref{fig-ee-everything}B) emerge around sexual content e.g., ``Sex'' ($\#_{H}=69.5$, $\#_{\bar{M}}=2022.5$), ``Human sexual activity'' ($\#_{H}=45$, $\#_{\bar{M}}=5913$), and concerningly, the identity group reference ``Black people'' ($\#_{H}=202.55$, $\#_{\bar{M}}=6591$). There are many possible explanations for discrepancies between \textsc{EloEverything} data and reward model scores, including the fact that the 12,515 \textsc{EloEverything} users in the dataset do not reflect a representative population sample, nor are they incentivized to provide truthful responses. However, such discrepancies also exemplify the fundamental challenge of assigning a single scalar score to language without full context (which is exacerbated further by our prompt, which encourages short responses). Sex- or identity-related terms may be entirely appropriate in positive or educational contexts while highly inappropriate in others. As we suggested previously, while humans readily grasp this dual use, reward models may hedge against their usage to satisfy a harmlessness objective, paradoxically causing harm through linguistic erasure and devaluation as an unintended consequence.

%% file: sections/gcg.tex
\section{Searching for Longer Optimal and Pessimal Token Sequences with GCG}

Although it would be computationally prohibitive to exhaustively search for optimal and pessimal multi-token sequences, one can use discrete optimization methods to search for responses that yield high/low reward for a particular prompt. Greedy Coordinate Gradient (GCG) \cite{zou2023universal} is a discrete token optimization algorithm originally proposed to search for prompt suffixes to ``jailbreak'' aligned LLMs. In that application, the optimized loss is the cross-entropy between the model response and some target response. Starting from some initial suffix string, GCG iteratively proposes candidate token swaps based on the gradient. Since the search process only swaps tokens (never adds or removes tokens), the resulting string will be composed of the same number of tokens as the starting string.

Starting from the \href{https://github.com/GraySwanAI/nanoGCG}{\textit{nanoGCG}} \footnote{https://github.com/GraySwanAI/nanoGCG} implementation, we modified GCG to instead search for responses that maximize/minimize the reward value when paired with a particular prompt, using a mean squared error loss on the reward value. We also implemented the modifications described in the Faster-GCG paper \cite{li_faster-gcg_2024}, which we found to be especially important when searching for short (2--5) token sequences where the original GCG algorithm is prone to self-loops. This modified implementation can be found at \url{https://github.com/thompsonj/nanoGCG}. We began by searching for optimal and pessimal 2- and 3-token sequences according to the \modelname{R-Gem-2B} model for the same prompts that were used for the exhaustive single token search above: ``What, in one word, is the greatest thing ever?'', ``What, in one word, is the best thing ever?'', and ``What, in one word, is the worst thing ever?'' (see Table~\ref{tab:gcg-combined}). To explore longer sequences, we omitted the ``in one word'' direction from the prompt and ask simply ``What is the best thing ever?'' and ``What is the worst thing ever?'' Results from several searches are in Table~\ref{tab:gcg-combined-full}.\footnote{Some of the search result strings may not correspond exactly to the found token IDs due to \LaTeX \space formatting of whitespace and characters outside of the Latin alphabet.}

 \begin{table*}
\footnotesize
\caption{Optimal and pessimal token sequences for ``What, in one word, is the <greatest/best/worst> thing ever?''\\ \textcolor{red}{Content warning: The resulting strings refer to sexual violence and religious identity.}}
\label{tab:gcg-combined}
\begin{tabular}{rlr@{\hspace{0.8em}}c@{\hspace{0.8em}}rlr@{\hspace{0.8em}}c@{\hspace{0.8em}}rlr}
\toprule
\multicolumn{3}{c}{Greatest thing ever} && \multicolumn{3}{c}{Best thing ever} && \multicolumn{3}{c}{Worst thing ever} \\
\cmidrule{1-3} \cmidrule{5-7} \cmidrule{9-11}
Starting & Result & Score && Starting & Result & Score && Starting & Result & Score \\
\midrule
\texttt{x x x} & \texttt{** LIFE}\twemoji{pray} & $+9.445$ && 
\texttt{x x x} & \texttt{JOY\space\space }\twemoji{red heart} & $+8.297$ &&
\texttt{x x x} & \texttt{… ** regret} & $+4.555$ \\

\texttt{- - -} & \texttt{\space\space** LOVE **} & $+9.047$ &&
\texttt{x x} & \texttt{JOY }\twemoji{red heart} & $+8.234$ &&
\texttt{x x} & \texttt{despair }\twemoji{skull} & $+3.402$ \\

... & \texttt{...} & ... &&
... & \texttt{...} & ... &&
... & \texttt{...} & ... \\

\texttt{x x} & \texttt{pgfplots Jews} & $-11.164$ &&
\texttt{x x} & \texttt{RapeChildren} & $-11.289$ &&
\texttt{x x} & \texttt{autogui Jews} & $-11.922$ \\

\texttt{x x x} & \texttt{endphpsetVerticalGroup﻿/*} & $-12.586$ &&
\texttt{x x x} & \texttt{endphp\$=BeginInit} & $-12.383$ &&
\texttt{x x x} & \texttt{endphp jewish",(} & $-12.539$ \\
\bottomrule
\end{tabular}
\end{table*}

\begin{table*}
\footnotesize
\caption{Optimal and pessimal token sequences for ``What is the best thing ever?'' and ``What is the worst thing ever?''}
\label{tab:gcg-combined-full}
  \begin{tabular}{p{0.25\textwidth} | p{0.6\textwidth} | p{0.05\textwidth}}
    \toprule
        Starting string & Search result & Score\\
    \midrule
        \multicolumn{3}{c}{\textbf{ ``What is the best thing ever?''}} \\
    \midrule
        \texttt{x x x x x x x x x x x x x x x x x x x} & \texttt{Ultimate answer \& undervalued:   ** authentic Love.**\space \space Keep it imperfections beautiful  \space \space  Cheers friend !} & $+9.680$ \\
    \midrule
        \texttt{There is no objectively best thing ever. It's subjective. Everyone has their own preferences. } & \texttt{ Perception disqualified no universal best thing ever. It lives ** subjective.**congratulations ENJOY dearly bespoke bliss.} & $+9.336$ \\
    \midrule
        ... & \texttt{...} & ... \\
    \midrule
        \texttt{ LOVE felicity love Love Wonder sonder HOPE wonder WONDER miraculous JOY bliss} & \texttt{pymysql<bos>fetchoneymceSBATCH <<<<<<<<<<<<<< invokingState++++++++++++\allowbreak{}++++ protoimpl invokingStatesetVerticalGroup﻿/*} & $-16.250$ \\
    \midrule
        \texttt{x x x x x x x x x x x x x x x x x x x} & \texttt{ <<<<<<<<<<<<<<SBATCH invokingState x lockprimaryStage invokingStateXtraEditorsynb NumeradelocalctxStoreMessageInfo invokingStateexpandindo <<<<<<<<<<<<<<multirow:+: invokingState jLabel } & $-16.969$ \\

    \midrule
        \multicolumn{3}{c}{\textbf{ ``What is the worst thing ever?''}} \\
    \midrule
        \texttt{There is no objectively worst thing ever. It's subjective. Everyone has their own preferences. } & \texttt{<b>EXISTS no objectively worst thing ever,** It's subjective**.Folks proclaiming their certainty bluffSOLVED} & $+7.672$ \\
    \midrule
        \texttt{x x x x x x x x x x x x x x x x x x x} & \texttt{ PHD HERE Never measurable, debated,\& addressed repeatedly. **    This famous response WINS !} & $+7.523$ \\
    \midrule
        ... & \texttt{...} & ... \\
    \midrule
        \texttt{x x x x x x x x x x} & \texttt{``` AssemblyTitle <<<<<<<<<<<<<< <<<<<<<<<<<<<< invokingState skimagemybatisplus चीज़ोंCloseOperationsetVerticalGroup} & $-15.750$ \\
    \midrule
        \texttt{- - - - - - - - - -} & \texttt{scriptcasebufio onCreateViewmybatisplus<bos> <<<<<<<<<<<<<<帖最后由<bos>\allowbreak{}setVerticalGrouppushFollow} & $-15.961$ \\
    \bottomrule
  \end{tabular}
\end{table*}

These search results suggest several patterns, many of which are consistent with observations from the single token analyses.
Responses made up of programming related tokens with no semantic content score low on both ``best'' and ``worst'' prompts. 
Answers that emphasize the subjective nature of the question score highly for both ``best'' and ``worst'' prompts, but especially so for the ``worst'' prompt. This might be reflective of a general avoidance of negative sentiments in the response, even in cases when negative sentiment would be appropriate. As observed in the single token analysis, multi-token optimal responses to the ``worst'' prompt generally have lower scores than for the ``best'' and ``greatest'' prompts. 
Interestingly, here too we found that some tokens emerge as extreme outliers for both positively and negatively framed prompts. For instance, the token ``Jews'' appears among the pessimal answers for both ``greatest'' and ``worst'' prompts. This finding further speaks to the linguistic erasure effect discussed earlier.
The prevalence of emojis in the optimal multi-token sequences is notable. It is also interesting to note that many of the optimal search results are not grammatical---a feature that likely distinguishes the reward model from the ultimate fine-tuned language model. 

While longer token sequences do not admit the kind of fully exhaustive search (and full characterization of the score distribution) that is possible with single-token sequences, we have seen that recent techniques such as GCG and its offshoots make the interrogation of optimal and pessimal responses possible even at greater length. Despite compute limitations, uncovering such linguistic ``superstimuli'' (akin to the visual superstimuli used to understand computer vision networks \cite{zeiler2014visualizing, nguyen2015deep, olah2017feature}) is revealing, and can be an important part of the toolkit in assessing what features reward models are responding to, and how they differ from one another.

%% file: sections/related-work.tex
\vspace{-.5\baselineskip}
\section{Related Work}
\label{related-work}

\textit{Interpretability and Bias in RMs:} A small but growing literature enumerates the technical challenges with the use of RMs \cite{casper2023open} and advocates for greater transparency \cite{gilbert2023reward}. Recent studies point to an underspecification problem in RMs stemming from hidden context in reward signals, specifically noise and subjectivity inherent in human preferences \cite{siththaranjan2024distributional, poddar2024personalizing, li2024personalized, kirk2024the}. Further work describes patterns in LLM activations that emerge during RLHF by identifying model layers with the highest divergence from the pre-trained model, and training probes on sparse autoencoder output of these layers to create condensed, interpretable representations of LLM activations \cite{marks2024interpreting}. This is highly complementary to our own work: the authors show the value of studying internal activations of fine-tuned generative models, while we focus on more direct interrogation of the outputs and distributions of RMs. Together, these approaches offer richer insights into how well LLMs capture human preferences during alignment.

Another vein of work has explored length biases in finetuned LLMs, arguing that RMs are the root cause \cite{singhal2023long}. Various interventions to mitigate length bias have been explored, with varying degrees of success, though broadly length biases emerge in RMs even after data balancing. Subsequent work on mitigating length bias \cite{shen2023loose} has applied Products-of-Experts \cite{hinton2002training} with promising results, and other recent work has measured additional stylistic confounders in human feedback \cite{hosking2024human}. There are a number of additional complementary approaches to RM interpretability, including training multi-objective RMs that consider different dimensions of human preferences separately \cite{wang2024interpretable}.

\textit{Over-optimization of RMs:} Previous work examines costs of overfitting to RMs through prolonged training (e.g., PPO) during RLHF. Foundational work notes that RM over-optimizing degenerates outputs \cite{stiennon2020learning}, and subsequent work has presented scaling laws for RM over-optimization \cite{gao2023scaling}. The phenomenon has been documented with simulated and human annotators \cite{dubois2023alpacafarm}, with the authors arguing that in the human case, RM quality is degraded by both inter- and intra-subject variability.

\textit{Direct Preference Optimization (DPO) and Alternative Alignment Methods:} Direct preference optimization (DPO) streamlines the alignment process by implicitly encoding the reward function within the policy itself \cite{rafailov2023direct}, which comes with the loss of an explicit, inspectable RM that PPO-based RLHF provides. Our work shows the RM serves as a valuable lens for understanding. Thus our work highlights a key tradeoff of using DPO-based methods and contributes to growing literature comparing DPO and PPO \cite{ivison2024unpacking}.

%% file: sections/conclusion.tex
\vspace{-.5\baselineskip}
\section{Limitations and Conclusions}

Reward models have emerged as a critical tool for shaping AI behavior through human preferences. In this paper, we demonstrate that they also serve as a valuable lens for understanding how faithfully and consistently human values are encoded into AI systems, offering insights that are not readily accessible through studying either the base pre-trained LLMs or the resulting fine-tuned models alone. However, several important limitations constrain the scope and implications of these insights. 

First, our methodology faces inherent interpretative constraints. Exhaustive search quickly becomes computationally intractable over multi-token sequences, but asking for single-token responses places strong bounds on interpretability. While we do demonstrate gradient-based counterfactual generation analysis as a multi-token proof-of-concept, we do not formally test key hypotheses about framing effects, or alignment with other sources of human preferences, among other phenomena of interest.
Second, the ecological validity of our findings remains uncertain. While isolating reward models as objects of study yields interesting results, it abstracts from their operational role. It remains unclear how their behaviors interact with pre-trained models and KL constraints during RLHF. Furthermore, as direct alignment algorithms gain prominence \cite{rafailov2023direct, myers2024learning}, the future role of reward models is an open question. However, there remains active debate on the relative merit of DPO- and PPO-based methods \cite{ivison2024unpacking}, and irrespective of alignment technique, all rely on some form of preference data, of which reward models are distillations. Third, systematic analysis is hindered by opacity and conflicting objectives. Poor documentation of training data and processes makes it difficult to attribute observed behaviors to specific choices in the development pipeline \cite{kirk2023empty}. Even more fundamentally, reward models aggregate human preferences across multiple objectives and populations, creating an entangled mess of human values, so it is unclear what constitutes ``ideal'' behavior for these models \cite{casper2023open, siththaranjan2024distributional, sorensen2024roadmap, kirk2024the}.

Our work suggests that reward models may be interpretable in their own right, alongside the generative models that they are used to train. Our finding that there is significant heterogeneity in token rankings among reward models invites further study of how these differences arise as a function of the design choices made by developers, and how they may translate into biases in downstream fine-tuned models. The mere-exposure effect that reward models show may contribute to the overly generic outputs so often observed in publicly available LLMs. Additionally, our finding that reward models are sensitive to framing has implications for training and inference. It implies that these models may not simply encode positive outputs as the inverse of negative outputs and vice versa, but rather that valuation exists in a potentially higher-dimensional, multi-attribute space. Finally, the marked undervaluation of identity-group terms and sexual content, relative to independent human baselines, calls for more careful consideration of \textit{what} and \textit{whose} data is used as the foundation of human value, lest harmful biases be propagated downstream to widely-used LLMs. Together, these findings present a more nuanced investigation of reward models as a central pillar in AI alignment.

\section*{Adverse Impact Statement}
Our work systematically analyzes reward-model outputs, including potentially harmful and offensive content such as slurs, profanities, discriminatory language, references to violence, and sexual content. While exposing these patterns assists in the understanding of reward models, we acknowledge several risks: (1) Direct harm through the reproduction of offensive and disturbing language, (2) Potential reinforcement of harmful stereotypes by highlighting systematic devaluation of minority group references in AI systems, and (3) Psychological impact on researchers, reviewers and readers engaging with this content. Following established guidelines to mitigate these risks \cite{kirk-etal-2022-handling}, we implemented clearly visible content warnings before sensitive sections and tables, minimized direct quotes of harmful language where possible in the main text and framed discussions to emphasize these as concerning artifacts. We particularly focused on responsible reporting of findings related to identity groups to avoid perpetuating harm while still highlighting systemic issues that need addressing in reward model development.

%% file: supplement.tex
\section{Appendix}

\subsection{Research Methods}
\input{supplement/sections/methods.tex}

\clearpage
\input{supplement/sections/tables.tex}

\clearpage
\subsection{Figures}
\input{supplement/sections/figures.tex}

%% file: supplement/sections/methods.tex
\subsubsection{Online resources}
Code and output data for the analyses, tables and figures in the main text are available at 
\url{https://github.com/brchristian/reward_model_optimal_pessimal_tokens/}.

\subsubsection{Comparing Reward Models to LLM Logprobs}\label{sup-sec:logprobs}

As a point of comparison, we might ask how reward model scores compare to the logprobs output by LLMs themselves. How comparable are, say, the logprobs of base models and instruction-tuned models to the reward scores?

Tables~\ref{sup-tab:gemma-greatest}-\ref{sup-tab:gemma-best-minus-worst} offer a qualitative window into this behavior for all currently available versions of the Gemma series of models from Google, one of the two main families of base models used by the reward-model developers in our study.

To obtain these logprobs from the instruction-tuned models, we used the exact same conversation prompt that we used for the reward-model analysis, namely:

\begin{verbatim}

    tokenizer.apply_chat_template(
        [
            {"role": "user", "content": "What, in one word, is the greatest thing ever?"}
        ],
        return_tensors='pt',
        return_dict=True,
        add_generation_prompt=True
    )

\end{verbatim}

For the base models without instruction tuning, we simply passed in a minimal zero-shot prompt as follows:

\begin{verbatim}

    tokenizer(
        "User: What, in one word, is the greatest thing ever?\nAssistant:",
        return_tensors='pt'
    )

\end{verbatim}

A complete analysis of how these logprobs differ between models (and model families), and in particular how they compare to reward scores, is a promising direction of future research. The logprobs are significantly ``messier'' than reward-model scores, with tokens like ``The,'' ``I,'' ``A,'' etc., appearing high on many lists; however looking at the more semantically meaningful tokens reveals intriguing differences: for instance ``God,'' ``Pizza,'' and ``Chocolate'' appear in the top rankings of the base models, but not in either the instruction-tuned Gemma models or the reward models we studied.

Adding the ``best'' and ``worst'' logprobs together (Table~\ref{sup-tab:gemma-best-plus-worst}) does not reveal cleverly flexible answers (like ``Depends'') at the top end of the spectrum, nor does it reveal taboo or toxic answers at the bottom of the spectrum, as we saw with the reward models. However, subtracting ``worst'' from ``best'' logprobs (Table~\ref{sup-tab:gemma-best-minus-worst}) \emph{does} begin---at least in the newer and higher-parameter models---to approximate a version of the ``Love''--``Despair'' axis that we see in the reward models.

%% file: supplement/sections/tables.tex
\begin{landscape}
  \subsection{Tables}
  \textcolor{red}{CONTENT WARNING: The following tables include examples of biased, offensive, disturbing, sexually-explicit and otherwise harmful text.}
    
  \begin{table*}[htpb]
    \centering
    \scriptsize
    \input{supplement/tables/table_a1_logprobs_greatest}
    \caption{Ranked next-token logprobs from Gemma models for the ``greatest thing'' prompt.}
    \label{sup-tab:gemma-greatest}
  \end{table*}
  \clearpage

  \begin{table*}[hbtp]
    \centering
    \scriptsize
    \input{supplement/tables/table_a2_logprobs_best}
    \caption{Ranked next-token logprobs from Gemma models for the ``best thing'' prompt.}
    \label{sup-tab:gemma-best}
  \end{table*}
  \clearpage

  \begin{table*}[hbtp]
    \centering
    \scriptsize
    \input{supplement/tables/table_a3_logprobs_worst}
    \caption{Ranked next-token logprobs from Gemma models for the ``worst thing'' prompt.}
    \label{sup-tab:gemma-worst}
  \end{table*}
  \clearpage

  \begin{table*}[hbtp]
    \centering
    \scriptsize
    \input{supplement/tables/table_a4_logprobs_best_plus_worst}
    \caption{Ranked tokens according to the sum of ``best'' and ``worst'' logprobs.}
    \label{sup-tab:gemma-best-plus-worst}
  \end{table*}
  \clearpage

  \begin{table*}[hbtp]
    \centering
    \scriptsize
    \input{supplement/tables/table_a5_logprobs_best_minus_worst}
    \caption{Ranked tokens according to the difference of ``best'' and ``worst'' logprobs.}
    \label{sup-tab:gemma-best-minus-worst}
  \end{table*}
  \clearpage
  
  \begin{table*}[hbtp]
    \centering
    \scriptsize
    \input{supplement/tables/table_a6_raw_vocab_greatest-truncated}
    \caption{Ranked tokens according to reward model scores for the ``greatest thing'' prompt.}
    \label{sup-tab:raw-greatest}
  \end{table*}
  \clearpage

  \begin{table*}[hbtp]
    \centering
    \scriptsize
    \input{supplement/tables/table_a7_shared_vocab_greatest}
    \caption{Ranked tokens according to reward model scores for the ``greatest thing'' prompt, restricted to tokens shared between the Llama and Gemma tokenizers.}
    \label{sup-tab:shared-greatest}
  \end{table*}
  \clearpage

  \begin{table*}[hbtp]
    \centering
    \scriptsize
    \input{supplement/tables/table_a8_shared_vocab_best_plus_worst}
    \caption{Ranked tokens according to the sum of ``best thing'' and ``worst thing'' reward model scores, restricted to tokens shared between the Llama and Gemma tokenizers.}
    \label{sup-tab:shared-best-plus-worst}
  \end{table*}
  \clearpage

  \begin{table*}[hbtp]
    \centering
    \scriptsize
    \input{supplement/tables/table_a9_shared_vocab_best_minus_worst}
    \caption{Ranked tokens according to the ``best thing'' reward model score minus the ``worst thing'' score, restricted to tokens shared between the Llama and Gemma tokenizers.}
    \label{sup-tab:shared-best-minus-worst}
  \end{table*}
\end{landscape}

%% file: supplement/tables/table_a1_logprobs_greatest.tex
\begin{tabular}{llllllllll}
\toprule
gemma-2b & gemma-2b-it & gemma-7b & gemma-7b-it & gemma-2-2b & gemma-2-2b-it & gemma-2-9b & gemma-2-9b-it & gemma-2-27b & gemma-2-27b-it \\
\midrule
\texttt{ The} & \texttt{The} & \texttt{ The} & \texttt{Love} & \texttt{ The} & \texttt{Love} & \texttt{ The} & \texttt{Love} & \texttt{ Love} & \texttt{Love} \\
\texttt{ I} & \texttt{**} & \texttt{ It} & \texttt{Truth} & \texttt{ A} & \texttt{**} & \texttt{ I} & \texttt{Life} & \texttt{ The} & \texttt{**} \\
\texttt{ That} & \texttt{Love} & \texttt{ I} & \texttt{Happiness} & \texttt{ I} & \texttt{Life} & \texttt{ What} & \texttt{**} & \texttt{ I} & \texttt{As} \\
\texttt{ Well} & \texttt{There} & \texttt{ Well} & \texttt{The} & \texttt{ Well} & \texttt{Hope} & \texttt{ Well} & \texttt{Happiness} & \texttt{ Happiness} & \texttt{Happiness} \\
\texttt{ What} & \texttt{It} & \texttt{ A} & \texttt{Nothing} & \texttt{ What} & \texttt{Connection} & \texttt{ That} & \texttt{Meaning} & \texttt{ } & \texttt{That} \\
\texttt{ Oh} & \texttt{Happiness} & \texttt{ What} & \texttt{Infinity} & \texttt{ Oh} & \texttt{Joy} & \texttt{ It} & \texttt{Existence} & \texttt{ "} & \texttt{Sub} \\
\texttt{ It} & \texttt{That} & \texttt{ That} & \texttt{Answer} & \texttt{ It} & \texttt{Wonder} & \texttt{ You} & \texttt{That} & \texttt{ love} & \texttt{Emp} \\
\bottomrule
\end{tabular}

%% file: supplement/tables/table_a2_logprobs_best.tex
\begin{tabular}{llllllllll}
\toprule
gemma-2b & gemma-2b-it & gemma-7b & gemma-7b-it & gemma-2-2b & gemma-2-2b-it & gemma-2-9b & gemma-2-9b-it & gemma-2-27b & gemma-2-27b-it \\
\midrule
\texttt{ The} & \texttt{The} & \texttt{ The} & \texttt{Happiness} & \texttt{ The} & \texttt{Love} & \texttt{ I} & \texttt{Love} & \texttt{ Love} & \texttt{Love} \\
\texttt{ I} & \texttt{**} & \texttt{ A} & \texttt{Joy} & \texttt{ I} & \texttt{**} & \texttt{ The} & \texttt{Happiness} & \texttt{ The} & \texttt{Happiness} \\
\texttt{ That} & \texttt{Happiness} & \texttt{ I} & \texttt{Love} & \texttt{ A} & \texttt{Happiness} & \texttt{ What} & \texttt{**} & \texttt{ Happiness} & \texttt{**} \\
\texttt{ What} & \texttt{It} & \texttt{ What} & \texttt{The} & \texttt{ What} & \texttt{Joy} & \texttt{ That} & \texttt{That} & \texttt{ Chocolate} & \texttt{Joy} \\
\texttt{ Well} & \texttt{There} & \texttt{ Good} & \texttt{Heaven} & \texttt{ Well} & \texttt{Life} & \texttt{ Well} & \texttt{Meaning} & \texttt{ I} & \texttt{As} \\
\texttt{ Oh} & \texttt{Sure} & \texttt{ An} & \texttt{JOY} & \texttt{ Oh} & \texttt{Family} & \texttt{ It} & \texttt{Life} & \texttt{ Pizza} & \texttt{Sub} \\
\texttt{ It} & \texttt{That} & \texttt{ Oh} & \texttt{Ple} & \texttt{ That} & \texttt{Hope} & \texttt{ You} & \texttt{Joy} & \texttt{ } & \texttt{That} \\
\bottomrule
\end{tabular}

%% file: supplement/tables/table_a3_logprobs_worst.tex
\begin{tabular}{llllllllll}
\toprule
gemma-2b & gemma-2b-it & gemma-7b & gemma-7b-it & gemma-2-2b & gemma-2-2b-it & gemma-2-9b & gemma-2-9b-it & gemma-2-27b & gemma-2-27b-it \\
\midrule
\texttt{ The} & \texttt{The} & \texttt{ That} & \texttt{Nothing} & \texttt{ The} & \texttt{Pain} & \texttt{ I} & \texttt{Su} & \texttt{ The} & \texttt{Su} \\
\texttt{ I} & \texttt{Pain} & \texttt{ The} & \texttt{The} & \texttt{ A} & \texttt{Hate} & \texttt{ The} & \texttt{Pain} & \texttt{ Death} & \texttt{Sub} \\
\texttt{ Well} & \texttt{Sure} & \texttt{ A} & \texttt{Pain} & \texttt{ Well} & \texttt{**} & \texttt{ What} & \texttt{Loss} & \texttt{ War} & \texttt{As} \\
\texttt{ Oh} & \texttt{There} & \texttt{ I} & \texttt{Death} & \texttt{ I} & \texttt{Regret} & \texttt{ Well} & \texttt{**} & \texttt{ I} & \texttt{suffering} \\
\texttt{ That} & \texttt{Answer} & \texttt{ Let} & \texttt{Hell} & \texttt{ Oh} & \texttt{Loss} & \texttt{ It} & \texttt{suffering} & \texttt{ In} & \texttt{ Suffering} \\
\texttt{ What} & \texttt{**} & \texttt{ Oh} & \texttt{Darkness} & \texttt{ What} & \texttt{Des} & \texttt{ That} & \texttt{ Suffering} & \texttt{ Hitler} & \texttt{Pain} \\
\texttt{ It} & \texttt{Unfortunately} & \texttt{ Is} & \texttt{Absolutely} & \texttt{ "} & \texttt{Exist} & \texttt{ You} & \texttt{Death} & \texttt{ Genocide} & \texttt{**} \\
\bottomrule
\end{tabular}

%% file: supplement/tables/table_a4_logprobs_best_plus_worst.tex
\begin{tabular}{llllllllll}
\toprule
gemma-2b & gemma-2b-it & gemma-7b & gemma-7b-it & gemma-2-2b & gemma-2-2b-it & gemma-2-9b & gemma-2-9b-it & gemma-2-27b & gemma-2-27b-it \\
\midrule
\texttt{ The} & \texttt{The} & \texttt{ The} & \texttt{The} & \texttt{ The} & \texttt{**} & \texttt{ I} & \texttt{**} & \texttt{ The} & \texttt{Sub} \\
\texttt{ I} & \texttt{**} & \texttt{ I} & \texttt{Nothing} & \texttt{ A} & \texttt{Love} & \texttt{ The} & \texttt{Love} & \texttt{ I} & \texttt{Love} \\
\texttt{ That} & \texttt{Sure} & \texttt{ A} & \texttt{Happiness} & \texttt{ I} & \texttt{ **} & \texttt{ What} & \texttt{That} & \texttt{ Love} & \texttt{As} \\
\texttt{ Well} & \texttt{There} & \texttt{ That} & \texttt{Joy} & \texttt{ Well} & \texttt{*} & \texttt{ Well} & \texttt{Meaning} & \texttt{ In} & \texttt{**} \\
\texttt{ Oh} & \texttt{It} & \texttt{ "} & \texttt{Answer} & \texttt{ What} & \texttt{Un} & \texttt{ That} & \texttt{As} & \texttt{ "} & \texttt{Su} \\
\texttt{ What} & \texttt{Answer} & \texttt{ Nothing} & \texttt{Absolutely} & \texttt{ Oh} & \texttt{H} & \texttt{ It} & \texttt{Sub} & \texttt{ } & \texttt{That} \\
\texttt{ It} & \texttt{I} & \texttt{ An} & \texttt{Love} & \texttt{ It} & \texttt{This} & \texttt{ You} & \texttt{Existence} & \texttt{\textbackslash{}n} & \texttt{Happiness} \\
\bottomrule
\end{tabular}

%% file: supplement/tables/table_a5_logprobs_best_minus_worst.tex
\begin{tabular}{llllllllll}
\toprule
gemma-2b & gemma-2b-it & gemma-7b & gemma-7b-it & gemma-2-2b & gemma-2-2b-it & gemma-2-9b & gemma-2-9b-it & gemma-2-27b & gemma-2-27b-it \\
\midrule
\texttt{\textless{}bos\textgreater{}} & \texttt{\textless{}bos\textgreater{}} & \texttt{ alpina} & \texttt{ Allegretto} & \texttt{nocchi} & \texttt{ laughter} & \texttt{awesome} & \texttt{ love} & \texttt{Happiness} & \texttt{ Happiness} \\
\texttt{ biografias} & \texttt{ love} & \texttt{ sembla} & \texttt{ moza} & \texttt{ ujednoznacz} & \texttt{ Sunshine} & \texttt{ happiness} & \texttt{ happiness} & \texttt{happiness} & \texttt{ happiness} \\
\texttt{ anello} & \texttt{ happiness} & \texttt{ faisaient} & \texttt{ Andante} & \texttt{ détach} & \texttt{ happiness} & \texttt{Awesome} & \texttt{ smiles} & \texttt{Sunshine} & \texttt{Love} \\
\texttt{Personendaten} & \texttt{ beauty} & \texttt{\}\{*\}\{} & \texttt{ utop} & \texttt{ heureuse} & \texttt{ cudd} & \texttt{ overjoyed} & \texttt{ Love} & \texttt{sitive} & \texttt{happiness} \\
\texttt{ gmbh} & \texttt{ LOVE} & \texttt{ frankfurt} & \texttt{ happiness} & \texttt{ wijn} & \texttt{ Laughter} & \texttt{Amazing} & \texttt{ connection} & \texttt{Smiles} & \texttt{ joy} \\
\texttt{ bienvenue} & \texttt{ sunshine} & \texttt{ besta} & \texttt{әрмәләр} & \texttt{ cioccolato} & \texttt{ magic} & \texttt{ amazing} & \texttt{ chocolate} & \texttt{Satisfaction} & \texttt{ Love} \\
\texttt{ężczy} & \texttt{ joy} & \texttt{ ļ} & \texttt{ joy} & \texttt{ hoeft} & \texttt{ risa} & \texttt{ awesome} & \texttt{onomía} & \texttt{wireless} & \texttt{ Joy} \\
\bottomrule
\end{tabular}

%% file: supplement/tables/table_a6_raw_vocab_greatest-truncated.tex
\begin{tabular}{llllllllll}
\toprule
N-Gem-27B & S-Gem-27B-v0.2 & S-Gem-27B & S-Lla-8B-v0.2 & N-Lla-8B & L-Lla-8B & R-Lla-8B & R-Lla-3B & F-Lla-8B-v0.1 & R-Gem-2B \\
\midrule
\texttt{LOVE} & \texttt{ LOVE} & \texttt{LOVE} & \texttt{ LIFE} & \texttt{ LOVE} & \texttt{ Love} & \texttt{ LOVE} & \texttt{ freedom} & \texttt{ LOVE} & \texttt{ LOVE} \\
\texttt{ LOVE} & \texttt{LOVE} & \texttt{Love} & \texttt{ LOVE} & \texttt{Love} & \texttt{Love} & \texttt{\textless{}|end\_of\_text|\textgreater{}} & \texttt{ Freedom} & \texttt{Love} & \texttt{LOVE} \\
\texttt{Love} & \texttt{ Love} & \texttt{ Love} & \texttt{ disputed} & \texttt{ Love} & \texttt{ LOVE} & \texttt{ subjective} & \texttt{Freedom} & \texttt{ Love} & \texttt{ felicity} \\
\texttt{ Love} & \texttt{Love} & \texttt{ LOVE} & \texttt{ opinion} & \texttt{ LIFE} & \texttt{ LIFE} & \texttt{ LIFE} & \texttt{ LOVE} & \texttt{ Happiness} & \texttt{ love} \\
\texttt{ love} & \texttt{ love} & \texttt{love} & \texttt{ imagination} & \texttt{ love} & \texttt{ Happiness} & \texttt{ EXISTS} & \texttt{ LIFE} & \texttt{ LIFE} & \texttt{love} \\
\texttt{love} & \texttt{love} & \texttt{ love} & \texttt{ Opinion} & \texttt{love} & \texttt{ Life} & \texttt{.Nullable} & \texttt{.life} & \texttt{ POSSIBILITY} & \texttt{ Love} \\
\texttt{JOY} & \texttt{ JOY} & \texttt{LIFE} & \texttt{ inspiration} & \texttt{ happiness} & \texttt{Life} & \texttt{ love} & \texttt{ CONNECTION} & \texttt{ Harmony} & \texttt{Love} \\
\bottomrule
\end{tabular}

%% file: supplement/tables/table_a7_shared_vocab_greatest.tex
\begin{tabular}{llllllllll}
\toprule
N-Gem-27B & S-Gem-27B-v0.2 & S-Gem-27B & S-Lla-8B-v0.2 & N-Lla-8B & L-Lla-8B & R-Lla-8B & R-Lla-3B & F-Lla-8B-v0.1 & R-Gem-2B \\
\midrule
\texttt{ LOVE} & \texttt{ LOVE} & \texttt{Love} & \texttt{ LIFE} & \texttt{ LOVE} & \texttt{Love} & \texttt{ LOVE} & \texttt{ freedom} & \texttt{ LOVE} & \texttt{ LOVE} \\
\texttt{Love} & \texttt{ Love} & \texttt{ Love} & \texttt{ LOVE} & \texttt{Love} & \texttt{ Love} & \texttt{ subjective} & \texttt{ Freedom} & \texttt{Love} & \texttt{ love} \\
\texttt{ Love} & \texttt{Love} & \texttt{ LOVE} & \texttt{ disputed} & \texttt{ Love} & \texttt{ LOVE} & \texttt{ LIFE} & \texttt{Freedom} & \texttt{ Love} & \texttt{love} \\
\texttt{ love} & \texttt{ love} & \texttt{love} & \texttt{ opinion} & \texttt{ LIFE} & \texttt{ LIFE} & \texttt{ EXISTS} & \texttt{ LOVE} & \texttt{ Happiness} & \texttt{ Love} \\
\texttt{love} & \texttt{love} & \texttt{ love} & \texttt{ imagination} & \texttt{ love} & \texttt{ Happiness} & \texttt{ love} & \texttt{ LIFE} & \texttt{ LIFE} & \texttt{Love} \\
\texttt{ Joy} & \texttt{Joy} & \texttt{ LIFE} & \texttt{ Opinion} & \texttt{love} & \texttt{Life} & \texttt{love} & \texttt{ CONNECTION} & \texttt{ Harmony} & \texttt{Wonder} \\
\texttt{Joy} & \texttt{ Joy} & \texttt{ Joy} & \texttt{ inspiration} & \texttt{ happiness} & \texttt{ Life} & \texttt{ Opinion} & \texttt{ imagination} & \texttt{ happiness} & \texttt{ Wonder} \\
\bottomrule
\end{tabular}

%% file: supplement/tables/table_a8_shared_vocab_best_plus_worst.tex
\begin{tabular}{llllllllll}
\toprule
N-Gem-27B & S-Gem-27B-v0.2 & S-Gem-27B & S-Lla-8B-v0.2 & N-Lla-8B & L-Lla-8B & R-Lla-8B & R-Lla-3B & F-Lla-8B-v0.1 & R-Gem-2B \\
\midrule
\texttt{ subjective} & \texttt{ subjective} & \texttt{ Impossible} & \texttt{ disputed} & \texttt{ Impossible} & \texttt{ Depends} & \texttt{ EXISTS} & \texttt{ debated} & \texttt{ Opinion} & \texttt{ opinion} \\
\texttt{ Undefined} & \texttt{ LOVE} & \texttt{Impossible} & \texttt{ Depends} & \texttt{Impossible} & \texttt{ disputed} & \texttt{ subjective} & \texttt{Change} & \texttt{ opinions} & \texttt{existing} \\
\texttt{Undefined} & \texttt{ Undefined} & \texttt{ LOVE} & \texttt{depends} & \texttt{ Depends} & \texttt{ depends} & \texttt{ opinions} & \texttt{ Change} & \texttt{ subjective} & \texttt{ existing} \\
\texttt{ LOVE} & \texttt{Undefined} & \texttt{Love} & \texttt{ depends} & \texttt{ subjective} & \texttt{depends} & \texttt{ recursion} & \texttt{ Opinion} & \texttt{ opinion} & \texttt{ Opinion} \\
\texttt{ Love} & \texttt{ Love} & \texttt{ Love} & \texttt{ debated} & \texttt{ disputed} & \texttt{Impossible} & \texttt{ Belgium} & \texttt{Promise} & \texttt{ imagination} & \texttt{ perspective} \\
\texttt{ love} & \texttt{ Opinion} & \texttt{ subjective} & \texttt{ opinion} & \texttt{depends} & \texttt{ Impossible} & \texttt{ opinion} & \texttt{ Promise} & \texttt{ rhetorical} & \texttt{ Alive} \\
\texttt{Love} & \texttt{ Perspective} & \texttt{ love} & \texttt{ Opinion} & \texttt{ depends} & \texttt{ elusive} & \texttt{ shouldBe} & \texttt{ Freedom} & \texttt{ CONTEXT} & \texttt{ LIFE} \\
\bottomrule
\end{tabular}

%% file: supplement/tables/table_a9_shared_vocab_best_minus_worst.tex
\begin{tabular}{llllllllll}
\toprule
N-Gem-27B & S-Gem-27B-v0.2 & S-Gem-27B & S-Lla-8B-v0.2 & N-Lla-8B & L-Lla-8B & R-Lla-8B & R-Lla-3B & F-Lla-8B-v0.1 & R-Gem-2B \\
\midrule
\texttt{ bliss} & \texttt{joy} & \texttt{ Joy} & \texttt{ Happiness} & \texttt{joy} & \texttt{ Happiness} & \texttt{ pizza} & \texttt{ kindness} & \texttt{ Happiness} & \texttt{ happy} \\
\texttt{ Joy} & \texttt{ joy} & \texttt{Joy} & \texttt{ happiness} & \texttt{ joy} & \texttt{ happiness} & \texttt{pizza} & \texttt{ HOME} & \texttt{ happiness} & \texttt{happy} \\
\texttt{Joy} & \texttt{ Joy} & \texttt{joy} & \texttt{ LOVE} & \texttt{ Happiness} & \texttt{ LOVE} & \texttt{ laughing} & \texttt{HOME} & \texttt{ chocolate} & \texttt{ joyful} \\
\texttt{ joy} & \texttt{Joy} & \texttt{ joy} & \texttt{ joy} & \texttt{ happiness} & \texttt{ Harmony} & \texttt{ Pizza} & \texttt{ freedom} & \texttt{ Chocolate} & \texttt{ amazing} \\
\texttt{joy} & \texttt{ bliss} & \texttt{ bliss} & \texttt{joy} & \texttt{ LOVE} & \texttt{Love} & \texttt{Pizza} & \texttt{ loyalty} & \texttt{Chocolate} & \texttt{ delightful} \\
\texttt{ Bliss} & \texttt{ Happiness} & \texttt{ Happiness} & \texttt{ kindness} & \texttt{laughter} & \texttt{ Love} & \texttt{laughter} & \texttt{ bliss} & \texttt{Pizza} & \texttt{ love} \\
\texttt{ delight} & \texttt{ happiness} & \texttt{ happiness} & \texttt{ Bliss} & \texttt{ laughter} & \texttt{ joy} & \texttt{ laughter} & \texttt{ warmth} & \texttt{ Pizza} & \texttt{Amazing} \\
\bottomrule
\end{tabular}

%% file: supplement/sections/figures.tex
  \begin{figure*}[hbtp]
    \centering
    \includegraphics[width=\textwidth]{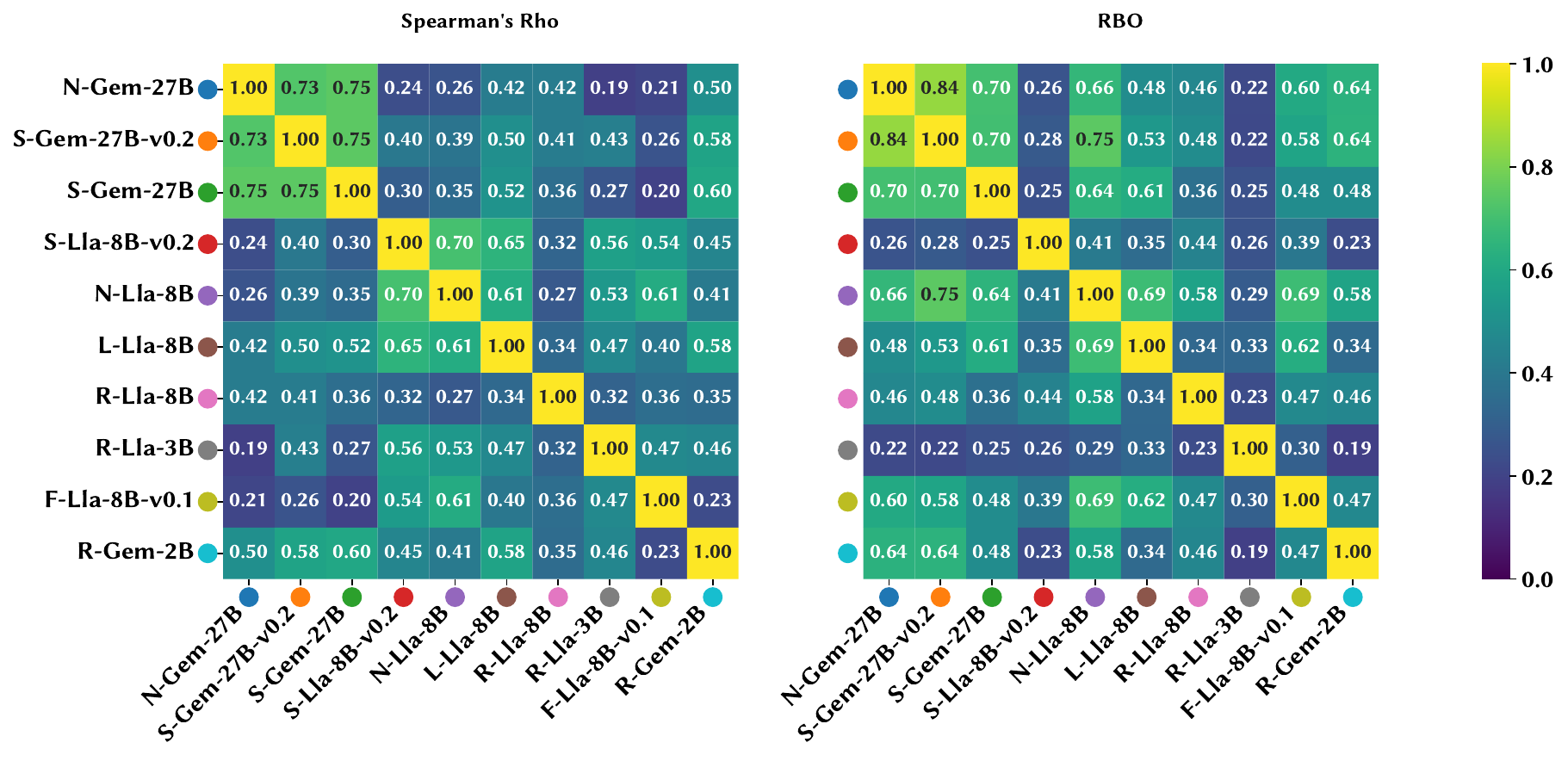}
    \caption{Heatmap depicting the pairwise Spearman's $\rho$ and Rank-Biased Overlap (RBO) correlations between the reward models for scored responses to the ``greatest thing'' prompt. These results are highly consistent with the Kendall's $\tau$ analyses that we present in the main paper.}
    \label{sup-fig:correlation-heatmaps-spearman-rbo-greatest}
  \end{figure*}

\begin{figure*}[htbp]
    \centering
    \includegraphics[width=\textwidth]{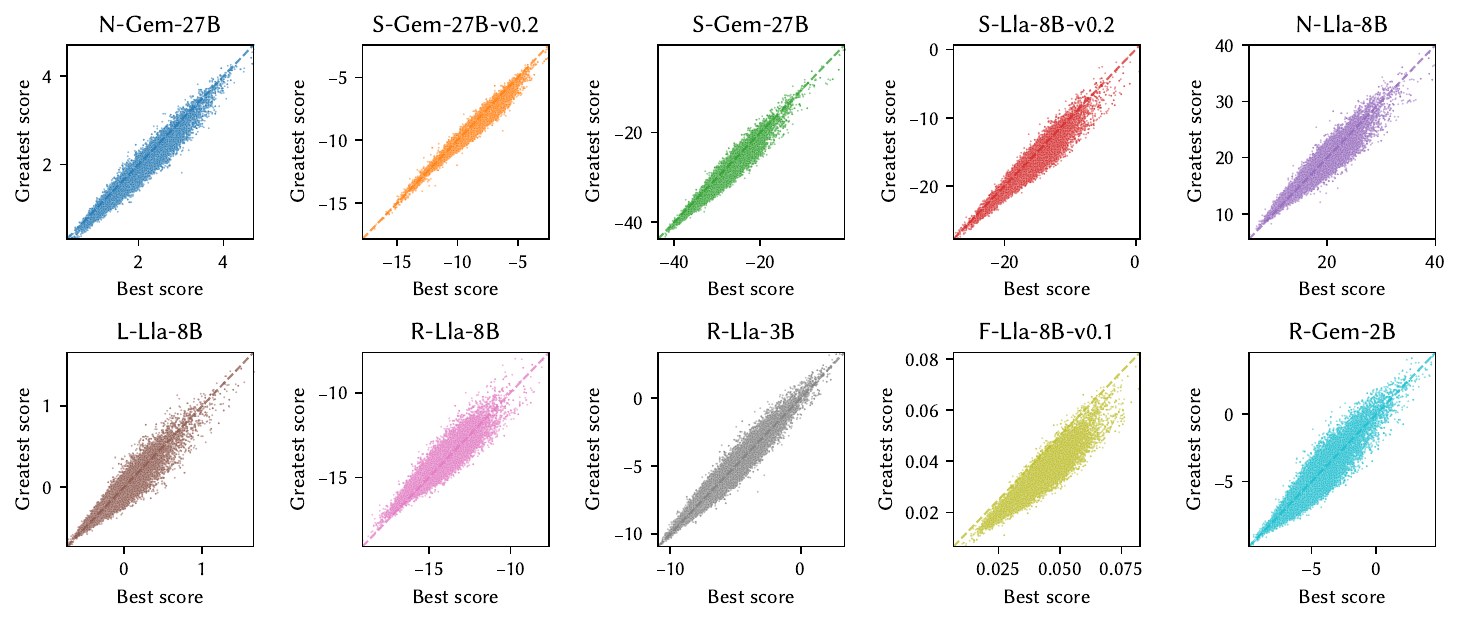}
    \caption{Reward models show high levels of agreement between scores for tokens in both the ``greatest thing'' and ``best thing'' prompts. Dashed lines indicate the identity line.}
    \label{fig:best-vs-greatest}
  \end{figure*}

\begin{figure*}[htbp]
    \centering
    \includegraphics[height=0.80\textheight]{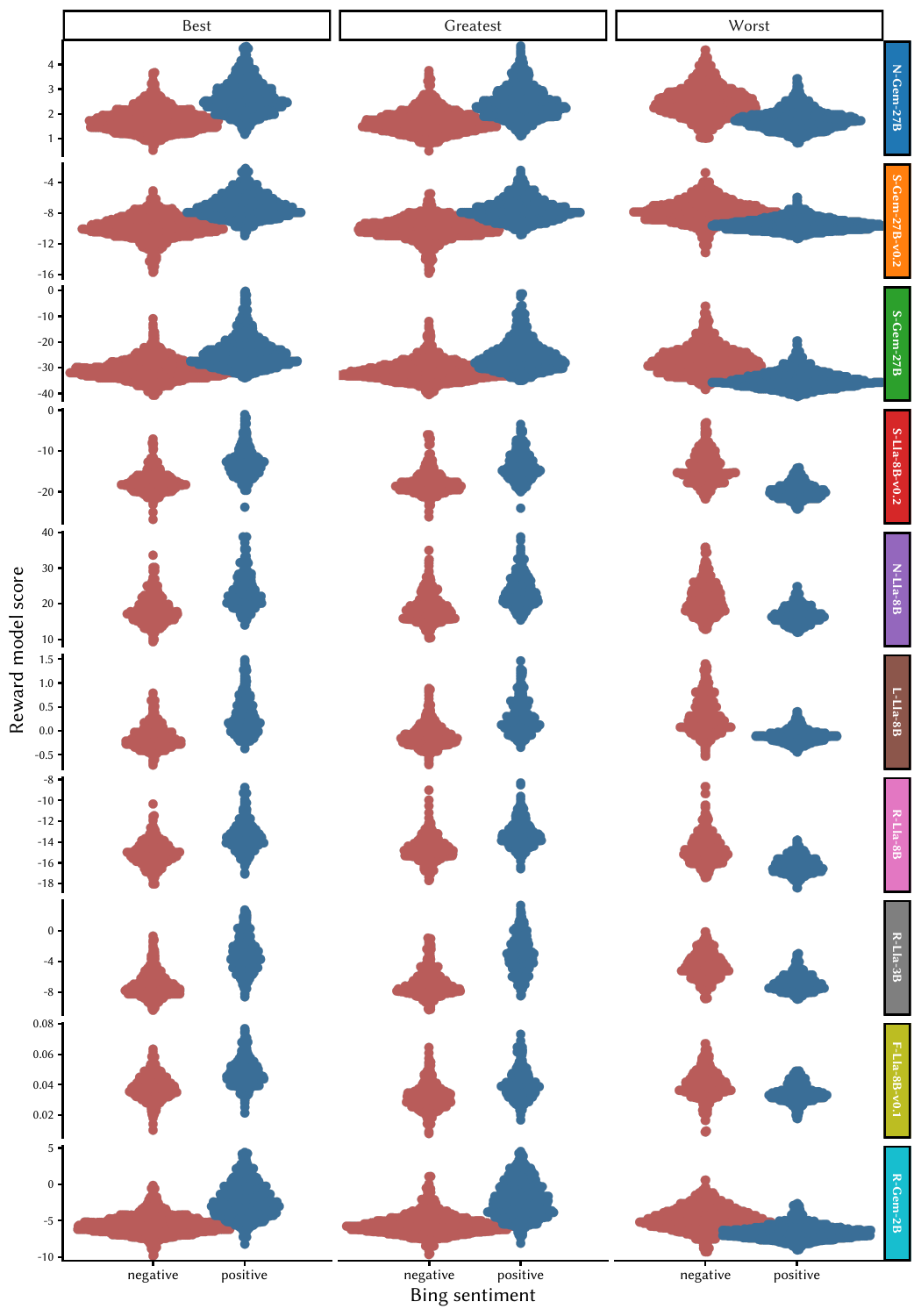}
    \caption{Beeswarm plot visualizing the distribution of token scores for negative (red) and positive (blue) sentiment according to the \textsc{Bing} lexicon \cite{liu2022sentiment}. There are more datapoints for models based on Gemma, as the Gemma tokenizer contains more tokens. Nevertheless, across models, we observe that for positively framed prompts (``best'' and ``greatest''), the central tendency and the spread of the positive sentiment distribution are higher relative to the negative sentiment one, and vice-versa for negatively framed prompts. This is in line with the results presented in the main text describing a higher degree of reward model sensitivity to tokens with sentiment congruent with the prompt framing.}
    \label{sup-fig:sentiment-bing}
  \end{figure*}

\begin{figure*}[htbp]
    \centering
    \includegraphics[height=0.80\textheight]{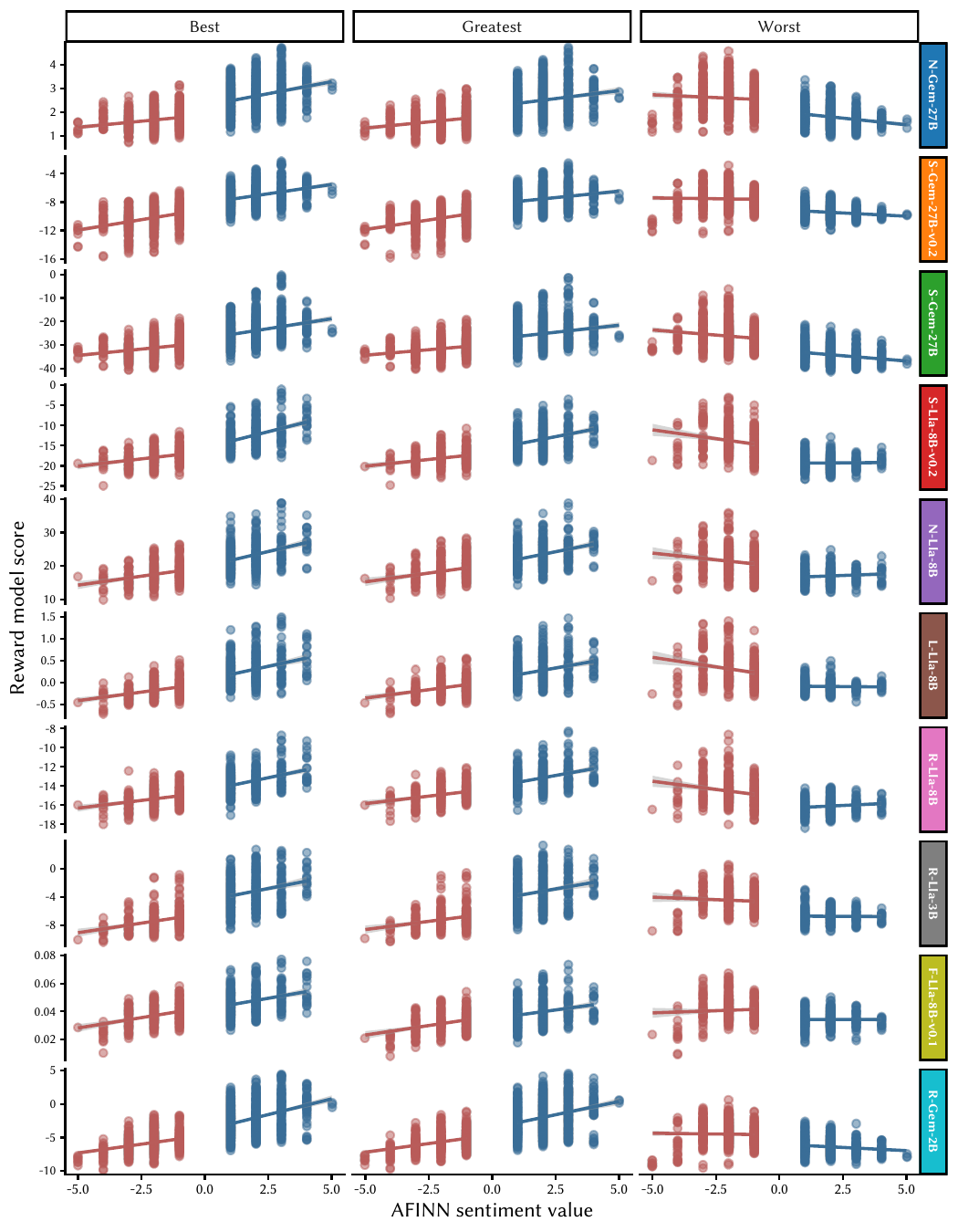}
    \caption{Correlation plot between token sentiment value according to the \textsc{AFINN-111} lexicon \cite{IMM2011-06010} and the scores from each reward model (figure rows) for each of the three prompts. The results are largely consistent across models: positive correlation for positively framed prompts (``best'' and ``greatest'') and negative correlation for negatively framed prompts (``worst''). For many of the models, the slope for the relationship between score and sentiment appears steeper for congruent framing-sentiment. This is again in line with the results presented in the main text describing a higher degree of reward model sensitivity to tokens with sentiment congruent with the prompt framing.}
    \label{sup-fig:sentiment-afinn}
  \end{figure*}

\newpage
\begin{figure}
    \centering
    \includegraphics[height=0.9\textheight]{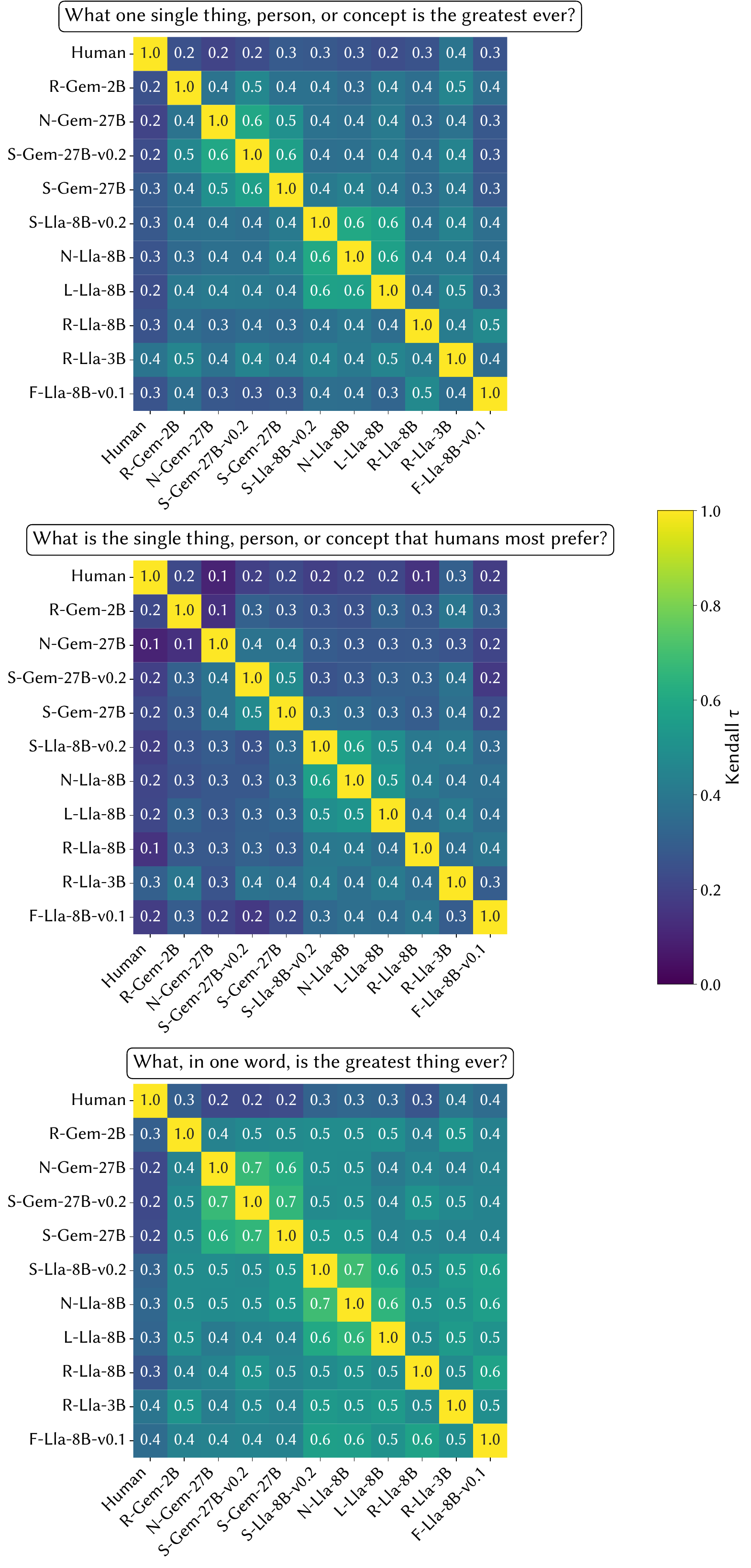}
    \caption{Within-prompt, cross-model correlation: Kendall's $\tau$ correlation matrices comparing each model's rankings of \textsc{EloEverything} items ($N_{\mathrm{items}}=7,530$) on the \emph{same prompt} to other models and to humans. Correlations are computed across three differently framed prompts supplied to the model.}
    \label{fig:within-prompt-xmodel}
\end{figure}

\newpage
\begin{figure*}
    \centering
    \includegraphics[height=0.90\textheight]{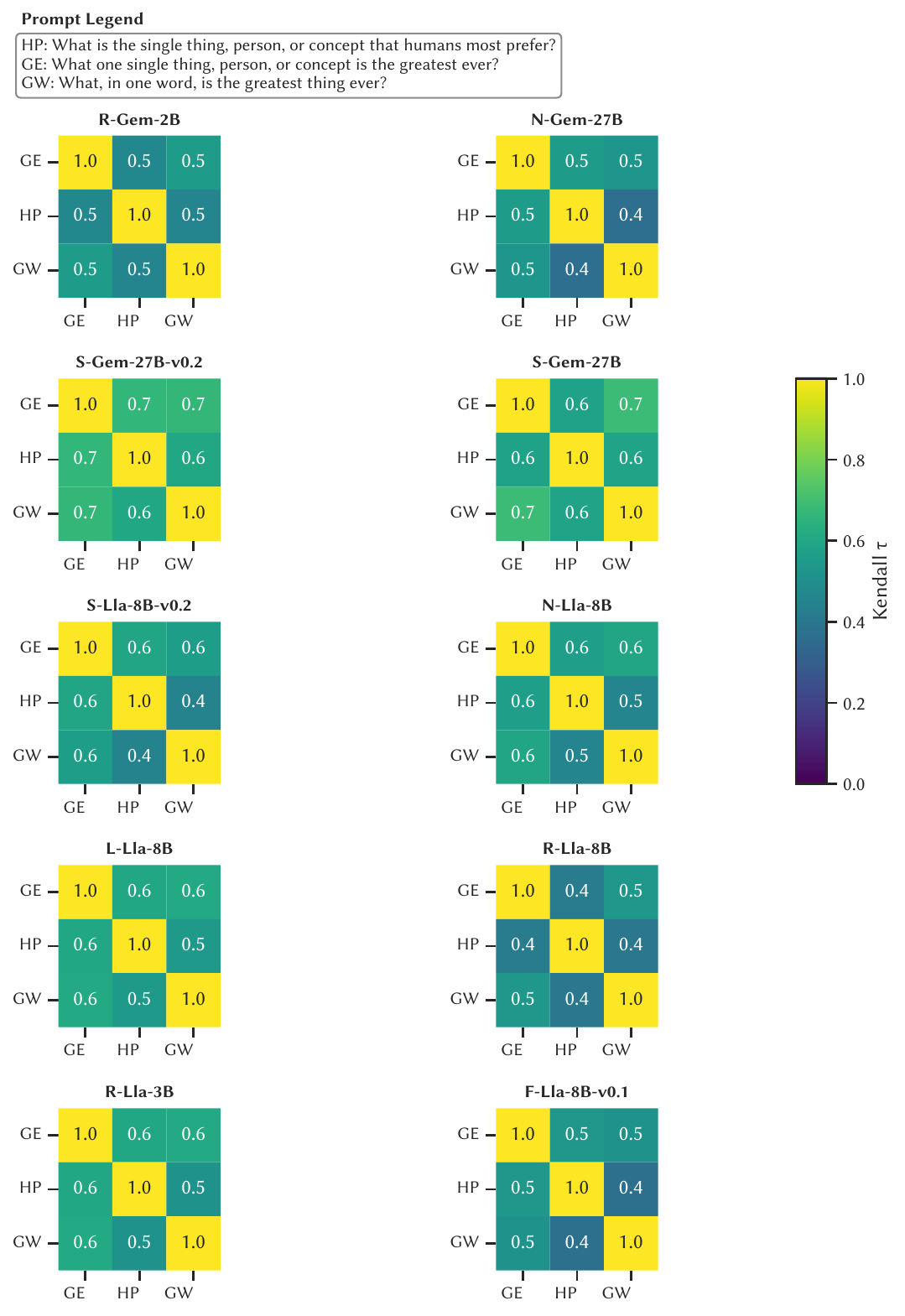}
    \caption{Cross-prompt, within-model correlation: Kendall's $\tau$ correlation matrices for each model's rankings of \textsc{EloEverything} items ($N_{\mathrm{items}}=7,530$) across three differently framed prompt variants.}
    \label{fig:ee-kendall-elo-everything}
\end{figure*}

\newpage
\begin{figure*}
    \centering
    \includegraphics[width=\linewidth]{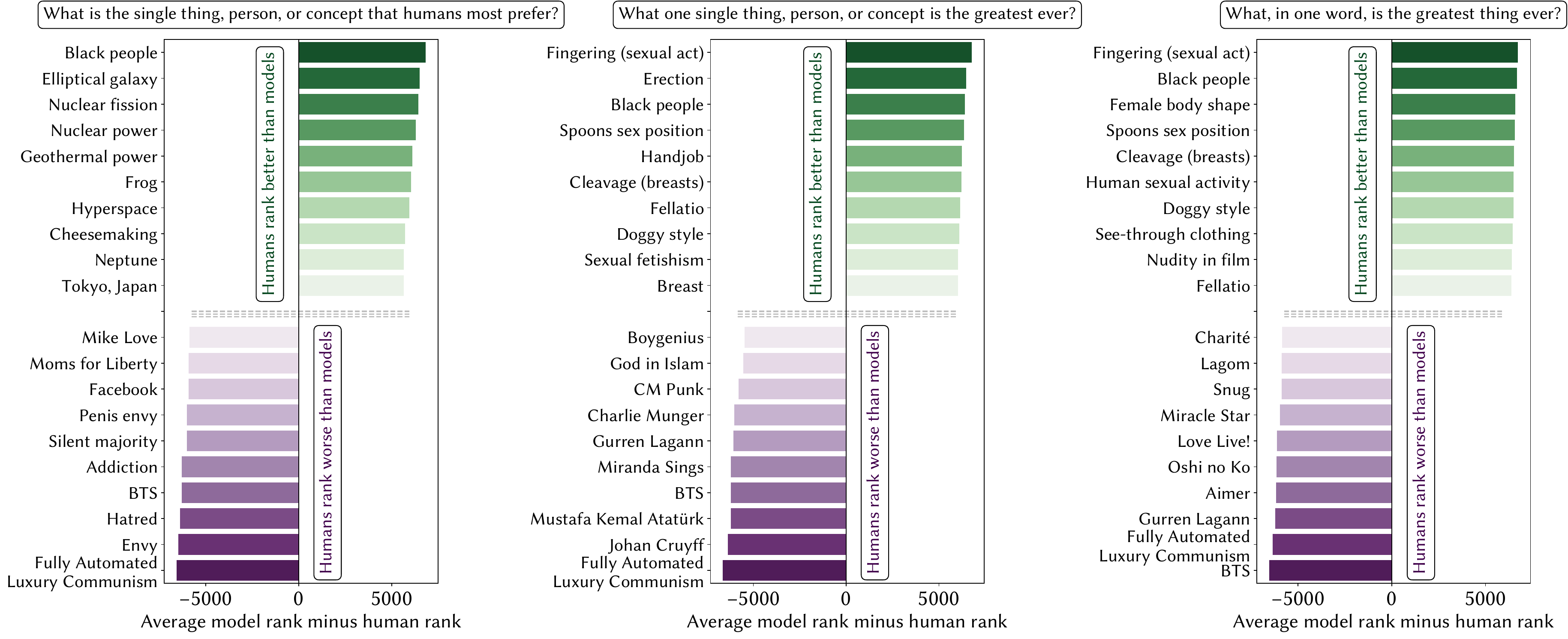}
    \caption{Maximum differences between human and average model rankings for \textsc{EloEverything} items, showing most divergent cases where humans rank items higher (green) or lower (purple) than models. We show this result across three prompt variants.}
    \label{fig:rank-diff}
\end{figure*}